\documentclass{article}
\usepackage{amssymb}

\usepackage[final]{corl_2025} % Uncomment for the camera-ready ``final'' version.
\usepackage{amsmath}
\usepackage{amsthm}
\usepackage{graphicx}
\usepackage{amsmath, amssymb}

\usepackage{mathtools}
\usepackage{soul}
\usepackage{algorithm}
\usepackage{subcaption}
\usepackage{mathrsfs}
\usepackage{amsmath,amsfonts,bm}

\def\eqref#1{(\ref{#1})}
\def\1{\bm{1}}

\DeclareMathAlphabet{\mathsfit}{\encodingdefault}{\sfdefault}{m}{sl}
\SetMathAlphabet{\mathsfit}{bold}{\encodingdefault}{\sfdefault}{bx}{n}

\usepackage[normalem]{ulem}
\theoremstyle{plain}

\theoremstyle{definition}

\theoremstyle{remark}

\newcommand{\bitem}{\begin{itemize}}
\newcommand{\eitem}{\end{itemize}}
\newcommand{\benum}{\begin{enumerate}}
\newcommand{\eenum}{\end{enumerate}}
\newcommand{\beq}{\begin{equation}}
\newcommand{\eeq}{\end{equation}}
\newcommand{\beqs}{\begin{equation*}}
\newcommand{\eeqs}{\end{equation*}}

\usepackage{amsfonts} %mathbb
\usepackage{algorithmic}

\usepackage[T1]{fontenc}    % use 8-bit T1 fonts
\usepackage[pagebackref=true]{hyperref}       % hyperlinks
\hypersetup{
    colorlinks=true,
    citecolor=black,
    linkcolor=blue,
    filecolor=magenta,      
    urlcolor=cyan,
    pdfpagemode=FullScreen,
    }
\renewcommand*\backref[1]{\ifx#1\relax \else (Cited on #1) \fi} % for citation
\usepackage{footnotebackref} % for footnote backref
\usepackage{lscape}
\usepackage{enumitem}

\usepackage{url}            % simple URL typesetting
\usepackage{booktabs}       % professional-quality tables
\usepackage{amsfonts}       % blackboard math symbols
\usepackage{nicefrac}       % compact symbols for 1/2, etc.
\usepackage{microtype}      % microtypography

\usepackage[addedmarkup=uline, defaultcolor=magenta, todonotes={textsize=scriptsize, textwidth=3.5cm}, authormarkuptext=name, commandnameprefix=always, xcolor]{changes} % just add a ``, final'' after xcolor as the last option to the changes package to make submittable version
\definechangesauthor[name={JD}, color=orange]{jd}

\newcommand{\ks}[1]{}

\newcommand{\toremove}[1]{}

\usepackage{wrapfig}
\usepackage{multirow}

\newcommand{\Regent}{\texttt{REGENT}}

\usepackage{bbm}

\usepackage{caption}
\usepackage{dsfont}
\newcommand{\RICL}{\texttt{RICL}}

\title{\RICL{}: 
Adding In-Context Adaptability to Pre-Trained Vision-Language-Action Models
}

\author{Kaustubh Sridhar$^1$, Souradeep Dutta$^{2}$, Dinesh Jayaraman$^1$, Insup Lee$^1$\\
$^1$University of Pennsylvania, $^2$University of British Columbia\\
\texttt{ksridhar@alumni.upenn.edu}
}

\begin{document}
\maketitle

%===============================================================================

\begin{abstract}
    % \jdtext{
    Multi-task ``vision-language-action'' (VLA) models have recently demonstrated increasing promise as generalist foundation models for robotics, achieving non-trivial performance out of the box on new tasks in new environments. However, for such models to be truly useful, an end user must have easy means to teach them to improve. For language and vision models, the emergent ability to perform in-context learning (ICL) has proven to be a versatile and highly useful interface to easily teach new tasks with no parameter finetuning. Unfortunately, VLAs pre-trained with imitation learning objectives do not naturally acquire ICL abilities. In this paper, we demonstrate that, with the right finetuning recipe and a small robot demonstration dataset, it is possible to inject in-context adaptability \textit{post hoc} into such a VLA. After retraining for in-context learning (\RICL{}), our system permits an end user to provide a small number (10-20) of demonstrations for a new task. \RICL{} then fetches the most relevant portions of those demonstrations into the VLA context to exploit ICL, performing the new task and boosting task performance. We apply \RICL{} to inject ICL into the $\pi_0$-FAST VLA, and show that it permits large in-context improvements for a variety of new manipulation tasks with only 20 demonstrations per task, without any parameter updates. When parameter updates on the target task demonstrations is possible, \RICL{} finetuning further boosts performance. We release code and model weights for \RICL{}-$\pi_0$-FAST alongside the paper to enable, for the first time, a simple in-context learning interface for new manipulation tasks\footnote{Website: \url{https://ricl-vla.github.io}}.
\end{abstract}

% Two or three meaningful keywords should be added here

%===============================================================================

\section{Introduction}

\begin{figure}[t]
    \centering
    \begin{subfigure}{\textwidth}
        \centering
        \fcolorbox{red}{white}{\includegraphics[width=0.35\linewidth]{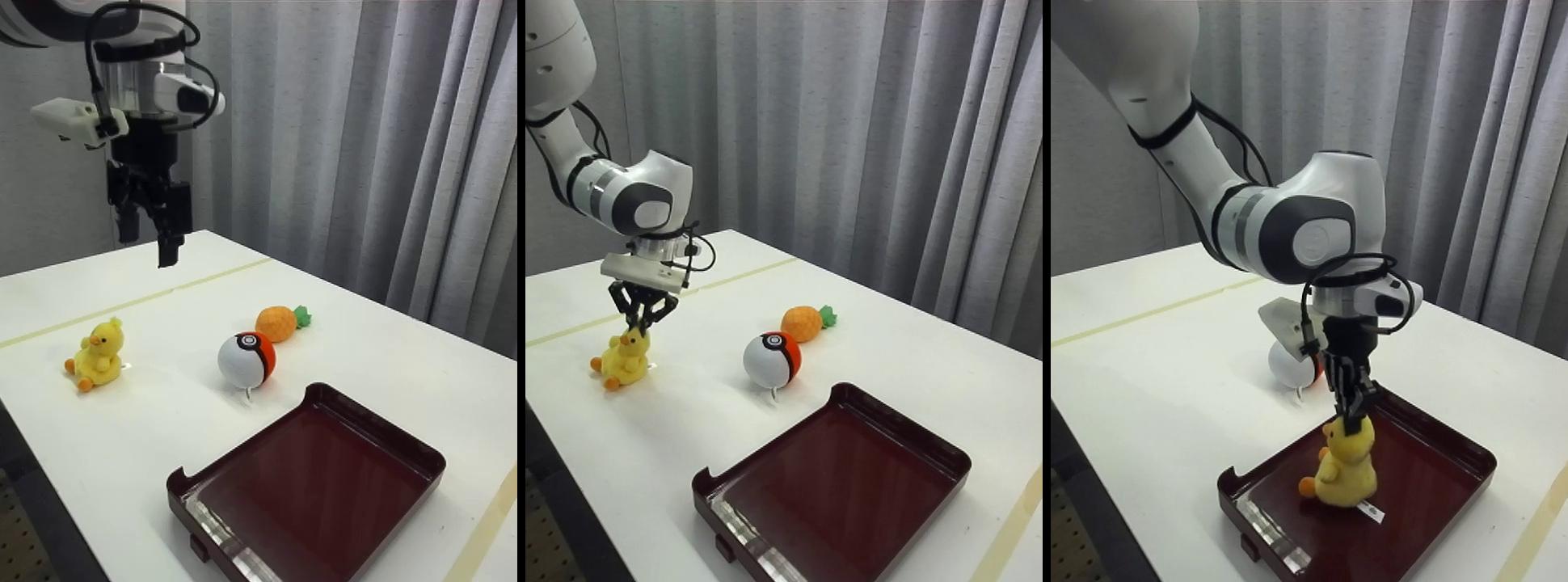}} \hspace{1em}
        \fcolorbox{green}{white}{\includegraphics[width=0.47\linewidth]{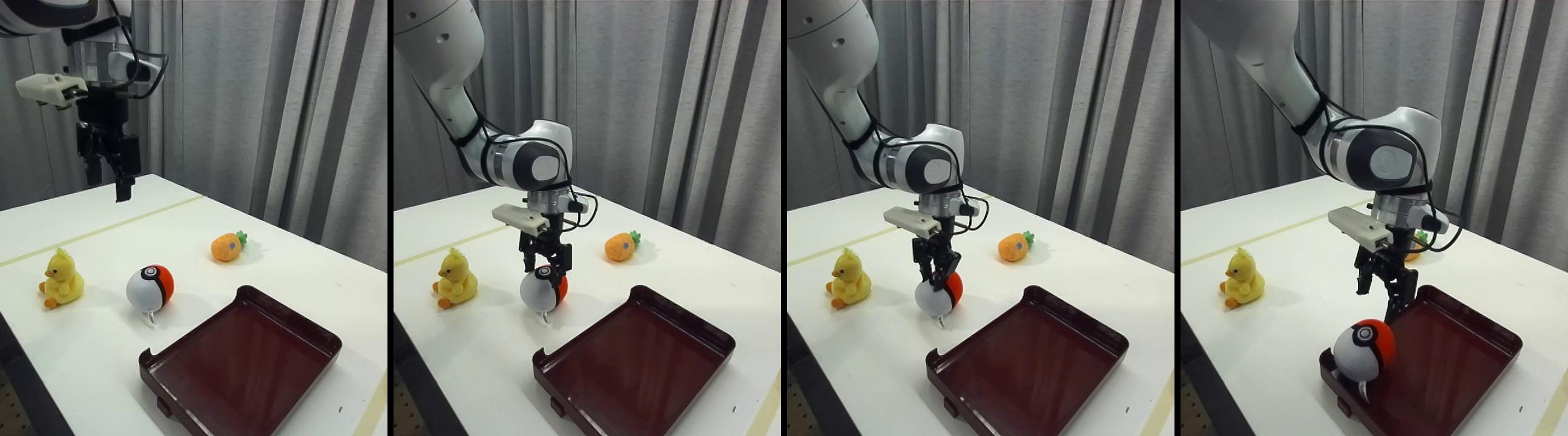}}
        \vspace{-0.5em}
        \caption{\scriptsize Task: "pick up the poke ball and put it in the tray". $\pi_0$-FAST-DROID \textbf{[L]} picks up the distractor (duck) instead (language grounding issue). \RICL{}-$\pi_0$-FAST-DROID \textbf{[R]} actually moves the unseen object (pokeball) with only RAG and ICL.}
    \end{subfigure}
    \begin{subfigure}{\textwidth}
        \centering
        \fcolorbox{red}{white}{\includegraphics[width=0.35\linewidth]{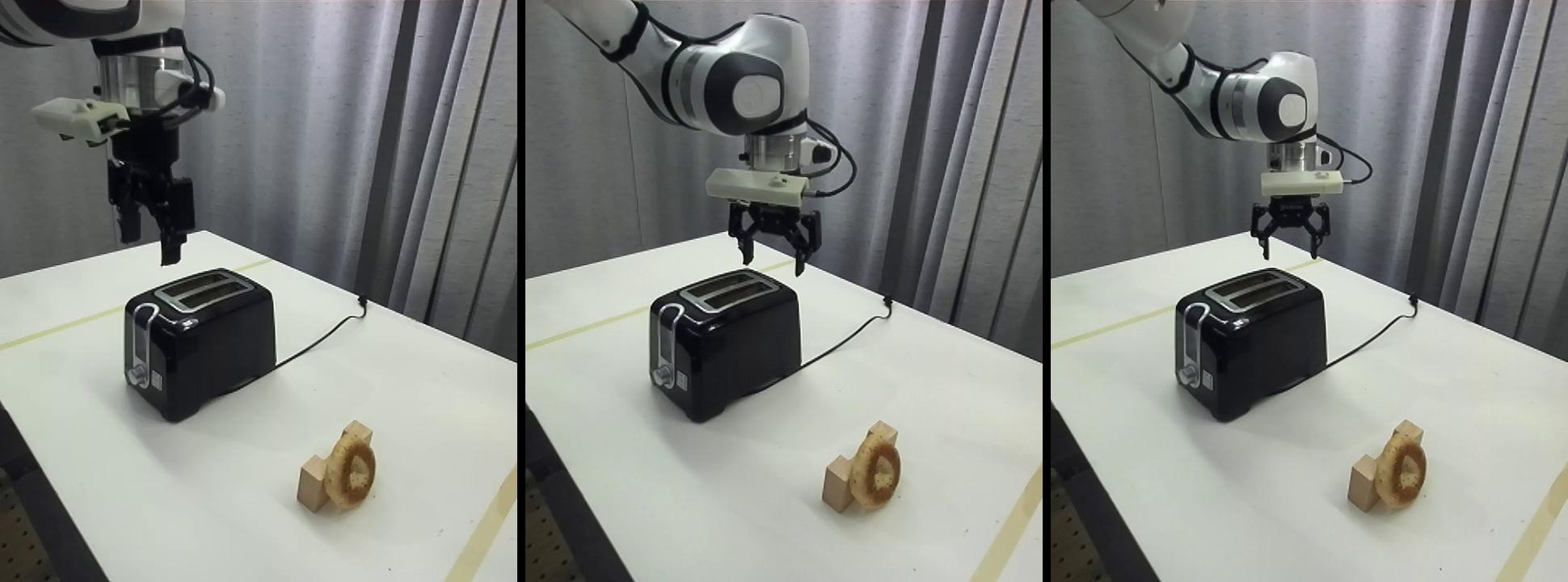}} \hspace{1em}
        \fcolorbox{green}{white}{\includegraphics[width=0.47\linewidth]{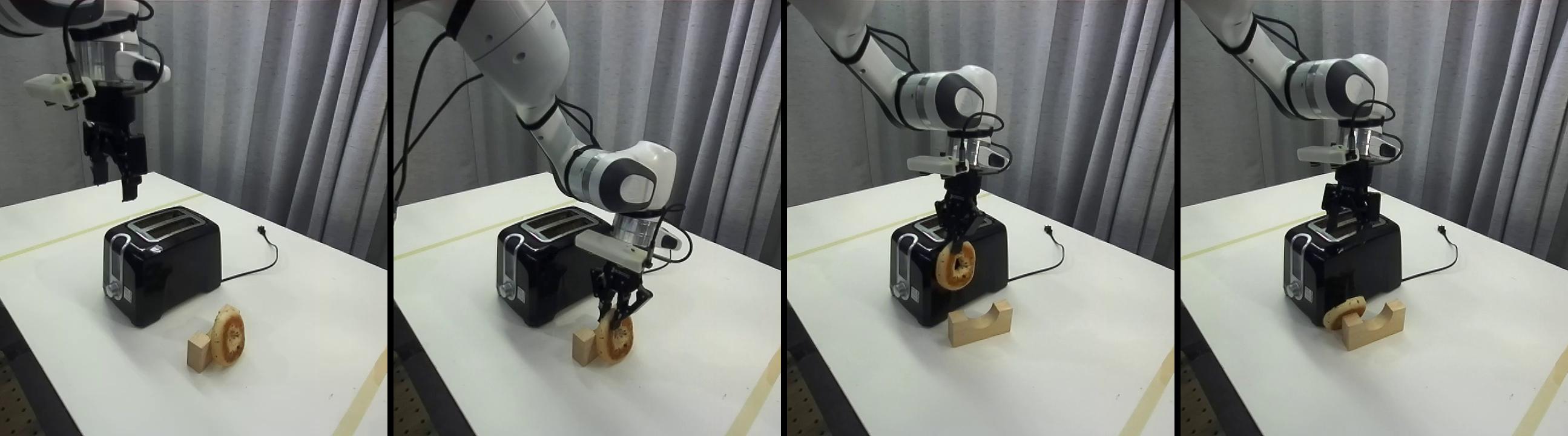}}
        \vspace{-0.5em}
        \caption{\scriptsize Task: "pick up the bagel and put it in the toaster". $\pi_0$-FAST-DROID \textbf{[L]} aimlessly wanders and cannot figure out the grasp or motion (adaptation issue). \RICL{}-$\pi_0$-FAST-DROID \textbf{[R]} almost completes the task (only with RAG and ICL) but drops the unseen object (bagel) at the end of the novel motion--a combination of an unfamiliar grasp at its rim, its unique initial vertical position, and the twist-and-lift motion.}
    \end{subfigure}
    \begin{subfigure}{\textwidth}
        \centering
        \fcolorbox{red}{white}{\includegraphics[width=0.35\linewidth]{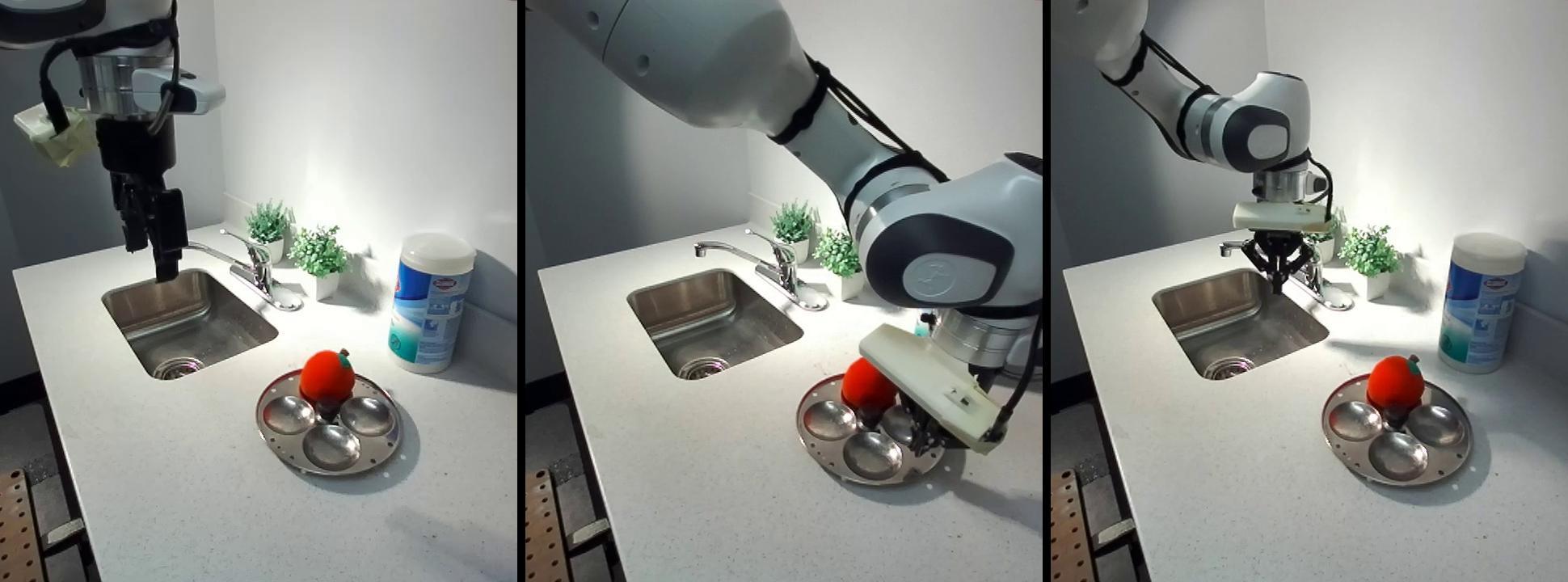}} \hspace{1em}
        \fcolorbox{green}{white}{\includegraphics[width=0.47\linewidth]{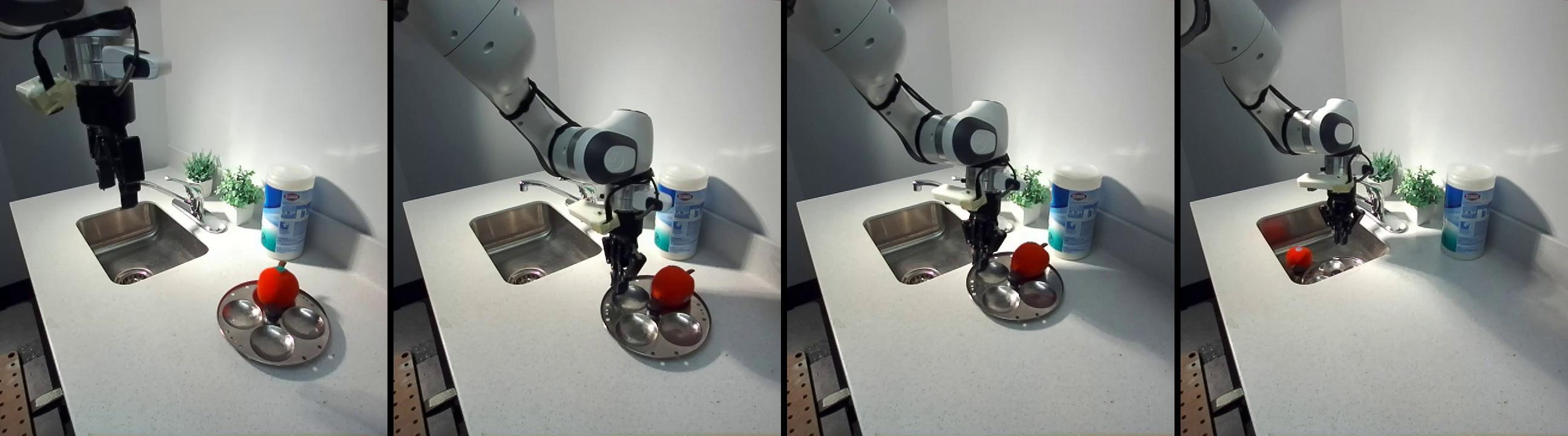}}
        \vspace{-0.5em}
        \caption{\scriptsize Task: "move the idli plate to the sink". In $\pi_0$-FAST-DROID's best test rollout shown here, it still struggles with the grasp and motion for this novel object (adaptation issue) or moves the apple (distractor) instead (language grounding issue). \RICL{}-$\pi_0$-FAST-DROID can perform the novel motion (gripper in depressions) on the unseen object (idli plate) in this new scene with new camera positions/orientations and with lighting changes (which is different from the table where the priming 
        % \jdc{careful with the ``training'' ... weren't you calling it priming or something?} 
        demonstrations were collected).}
    \end{subfigure}
    \begin{subfigure}{\textwidth}
        \centering
        \fcolorbox{red}{white}{\includegraphics[width=0.35\linewidth]{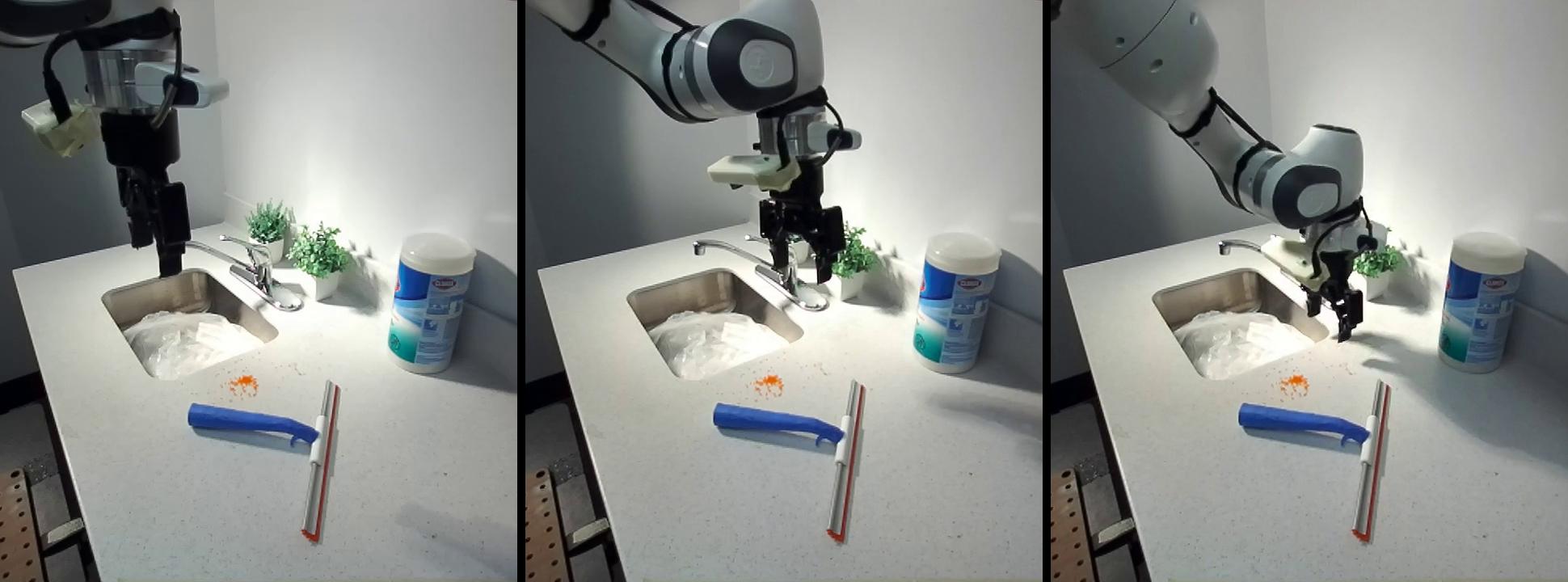}} \hspace{1em}
        \fcolorbox{green}{white}{\includegraphics[width=0.47\linewidth]{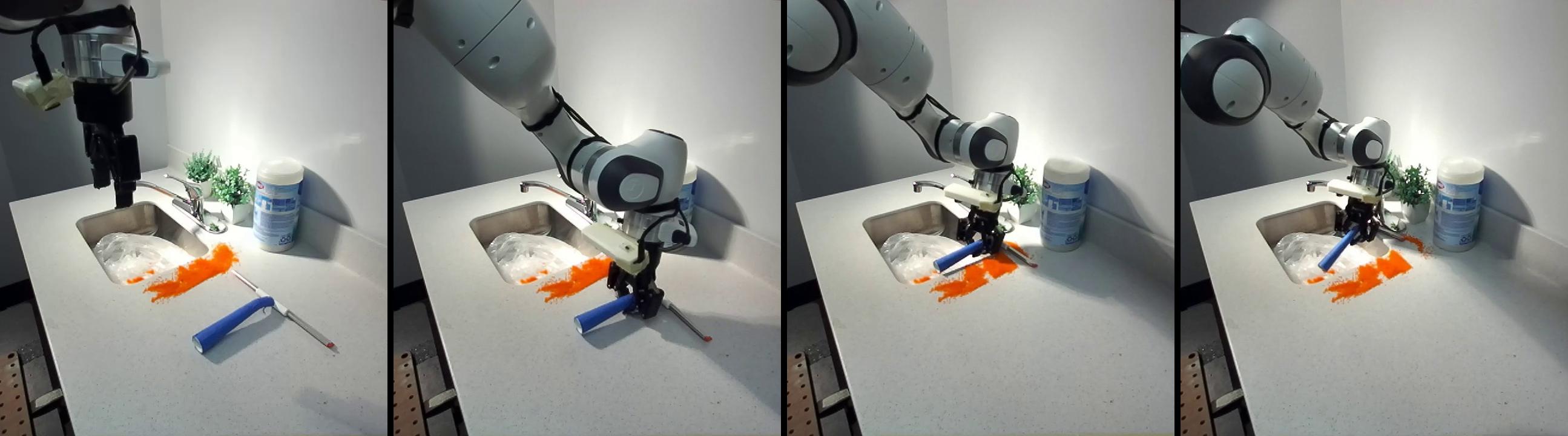}}
        \vspace{-0.5em}
        \caption{\scriptsize Task: "use the squeegee to clean the counter". $\pi_0$-FAST-DROID oscillates without success. It gets close, but it cannot figure out the grasp or the motion (adaptation issue). \RICL{}-$\pi_0$-FAST-DROID adapts to novel object (squeegee) and motion (part lifting, part dragging) in the new scene. Notice the pellets dropping into sink showing contact with the surface.}
    \end{subfigure}
    \begin{subfigure}{\textwidth}
        \centering
        \fcolorbox{red}{white}{\includegraphics[width=0.35\linewidth]{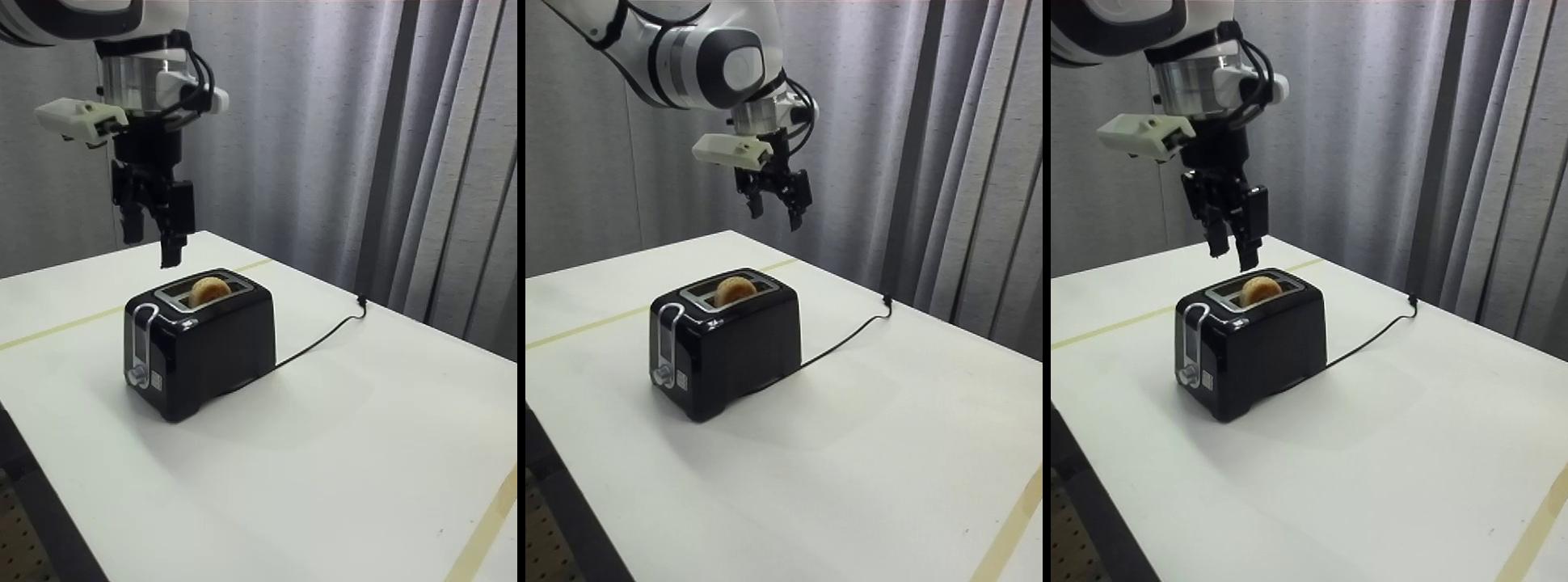}} \hspace{1em}
        \fcolorbox{green}{white}{\includegraphics[width=0.47\linewidth]{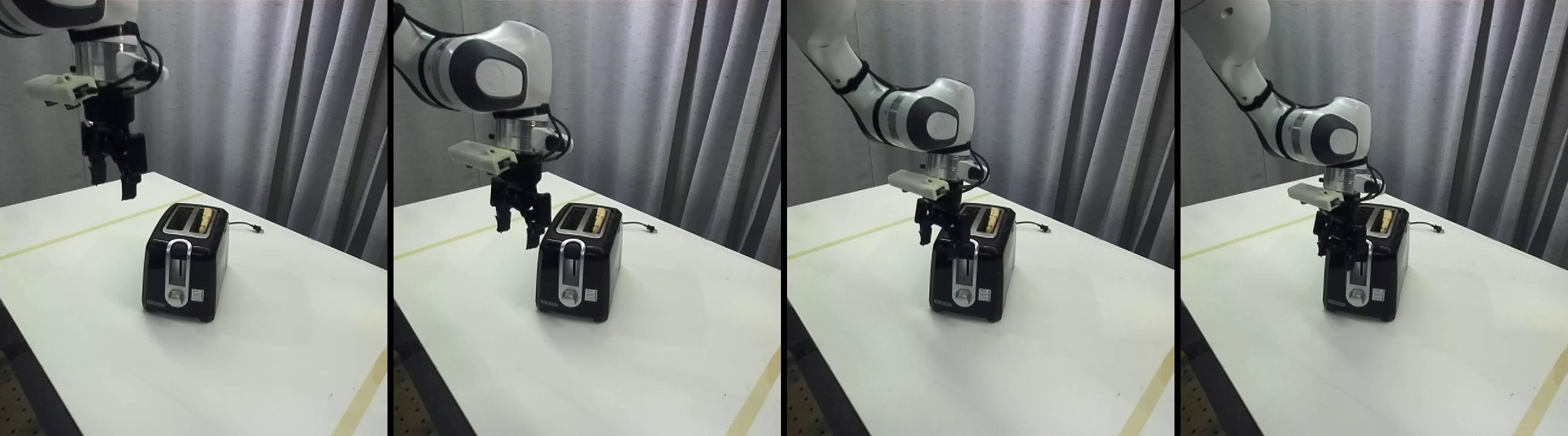}}
        \vspace{-0.5em}
        \caption{\scriptsize Task: "push the lever on the toaster". $\pi_0$-FAST-DROID \textbf{[L]} aimlessly wanders. It cannot figure out the precise location or the grasp (adaptation to a variant of training object issue). \RICL{}-$\pi_0$-FAST-DROID \textbf{[R]} completes the precise task, only with RAG and ICL, and with elicited latent actions not in the retrieval data (more information in Section \ref{sec:experiments_regentic_tuning}). This long-tail task appears infrequently in the DROID dataset.}
    \end{subfigure}
    \begin{subfigure}{\textwidth}
        \centering
        \fcolorbox{red}{white}{\includegraphics[width=0.35\linewidth]{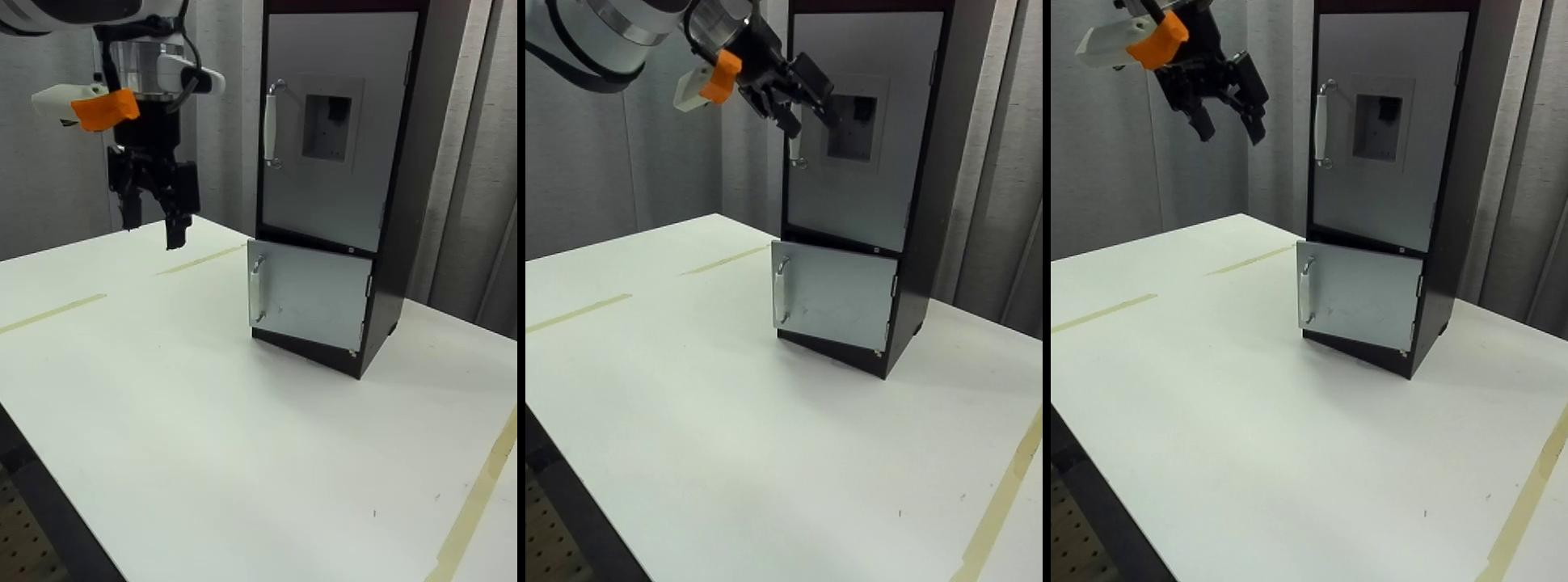}} \hspace{1em}
        \fcolorbox{green}{white}{\includegraphics[width=0.47\linewidth]{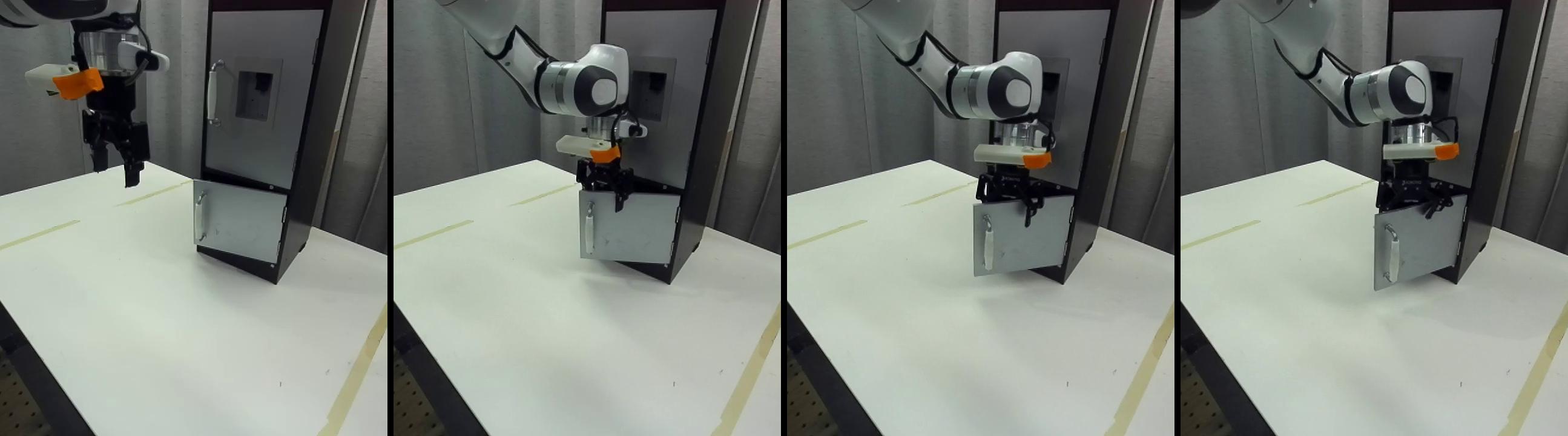}}
        \vspace{-0.5em}
        \caption{\scriptsize Task: "open the door of the bottom shelf". $\pi_0$-FAST-DROID \textbf{[L]} aimlessly wanders. It cannot figure out the motion to adjust or avoid the top door acting as obstacle (adaptation issue). \RICL{}-$\pi_0$-FAST-DROID \textbf{[R]} completes the task, with this variant of a seen object (this particular shelf) and novel motion (precise door opening adjusting for the obtructing top shelf), only with RAG+ICL. This is also a long-tail task.}
    \end{subfigure}
    \caption{\small Qualitative comparison between $\pi_0$-FAST-DROID \textbf{[L]} and \RICL{}-$\pi_0$-FAST-DROID \textbf{[R]}, with 20 task specific demonstrations for RAG and ICL, on new tasks, including novel objects, motions, and scenes. Additional comparisons can be found in Figure \ref{fig:qualitative_results_appendix} in Appendix \ref{app:additional_results_regentic_tuning}.
    % \jdc{how many examples did we use for teaching each task?} \ksc{20, why?} Unseen objects and novel motions are not present in the priming demonstrations  used to create \RICL{}-$\pi_0$-FAST-DROID or the DROID dataset \citep{droid} used before that to create $\pi_0$-FAST-DROID. \jdc{Seems very verbose for a caption. Can we simply say unseen objects and motions, or just ``unusual'' objects and motions? The main text can explain exactly what you mean by that.} \jdc{This figure would be a lot easier to parse (and also informative about your performance metrics) if you displayed on each frame how many of the subtasks you have now completed. OR at least show the score like ``2/3'' on the last frame alone for each method in each task. And you could color code it to quickly see how good each method is.} \jdc{Could you draw a grouping box around the base VLA frames and another around the RICL frames for each task? Would make it easier to parse. Could also try making them different colors. } Additional tasks' examples  can be found in Figure \ref{fig:qualitative_results_appendix} in the Appendix. \jdc{Not a fan of the size + placement of this figure on page 2. If it is here, it should be just a teaser. }
    }
    \label{fig:qualitative_results}
    \vspace{-1em}
\end{figure}
Robot learning is undergoing a transformative moment with the emergence of 
the first generation of
general-purpose Vision-Language-Action (VLA) models, capable of performing a wide spectrum of robotic tasks — a development with profound 
% and game-changing
practical implications. Such models \citep{pi0_fast, pi0_diffusion, OpenVLA, Octo, robocat, RT2, RTX, gemini_robotics} could address persistent challenges in robotics, including data scarcity, robustness, and generalization. 

A natural point of comparison for these VLAs is large language models (LLMs). 
%LLMs pre-trained on internet-scale datasets often outperform narrowly focused solutions and are already commonly used in our day-to-day lives. 
%used commonly today. 
%An important capability in LLMs 
%have demonstrated the ability 
%is that they 
One important factor in the widespread adoption of LLMs today is that they appear to be able to quickly learn new tasks, simply through providing a few examples as ``context'' alongside the query, with no parameter tuning. 
%not just via finetuning on task-specific data, but also simply via providing a few examples in their context. 
This capability, called in-context learning (ICL) \citep{in-context-learning}, emerges naturally in LLMs pre-trained for next-token prediction, due to the nature of web text data. Even better, one need not even manually provide these few examples. Instead, a retrieval mechanism could automatically fetch the most relevant data from a large corpus and place them into the LLM context. This retrieval-augmented generation (RAG) mechanism is widely adopted as a versatile interface to improve a base LLM \citep{rag_survey}. 
%These few-shot examples can be retrieved from task-specific datasets and this combination of retrieval-augmented generation (RAG) and in-context learning enables LLMs to more quickly and easily adapt compared to finetuning. 

Unfortunately, VLAs are trained with imitation learning objectives on relatively narrow demonstration datasets. % do not 
%on the other hand, do not seem to 
As one would expect, this does not 
naturally produce any
%naturally acquire 
in-context learning abilities. This means that ``improving'' a pre-trained VLA today means tuning its parameters on a new demonstration dataset \citep{pi0_diffusion}.  
% \jd{cite the original pi-zero  / gemini robotics papers both of which I think had some examples of these?}   % with imitation learning.
% \jd{I think this is missing one step in the logic of the abstract: LLMs naturally acquire in-context learning abilities through next-token prediction on web-scale data, but this is not true for VLA models trained with imitation learning ...}
%In this paper, we attempt to bring in-context learning and retrieval-augmentation to VLAs. 
%Such capabilities will enable quicker and easier adaptation and remove the additional burden of finetuning placed on the end user, especially when dealing with the real world. 
%A user would simply have to collect a few demonstrations of the robot performing a new task and provide that to the VLA system. This system would simply use RAG and in-context learning to perform this new task without any cumbersome training process. The versatility and ease of the user interface for teaching a new task would be unparalleled.
To make it possible for an end user to easily improve a VLA with no parameter tuning, we ask the following question: %Hence, in this paper, we answer the following question:
\begin{center}\textit{
How can we inject in-context learning abilities into a pre-trained VLA?
%quickly 
%convert a VLA into one that can adapt in-context to unseen tasks or the long tail of seen tasks?
}\end{center}
Once this is done, we should be able to painlessly boost the VLA's performance on any task, including handling unseen objects, novel motions, and new scenes that don't exist in the VLA training data. %, or even simply boost its performance on other . 

Our solution is to \textit{\textbf{r}etrain} for \textbf{i}n-\textbf{c}ontext \textbf{l}earning (\RICL{}, pronounced ``rickle''). \RICL{} borrows from prior recipes~\citep{PICL,kaustubh_regent_iclr} to train generalist models for in-context learning and RAG. In particular,
while REGENT~\citep{kaustubh_regent_iclr} trained generalist game-playing agents from scratch, \RICL{} 
uses this approach
%it repurposes the recently developed REGENT~\citep{kaustubh_regent_iclr} approach: while the original approach aimed to train a generalist game-playing agent from scratch, \RICL{}  uses the REGENT approach 
to instead post-train
% \jd{? mid-train? Is pre-train mid-train post-train etc an appropriate language to borrow to explain what we do here?} \ks{yes, post-train is fine}
an off-the-shelf VLA priming it to use its context effectively. The resulting \RICL{}-VLA
%generating a \RICL{}-VLA that can use it s co
%We propose \textbf{R}e-training for \textbf{I}n-\textbf{C}ontext \textbf{L}earning, \RICL{} (pronounced "rickle"), a method to 
%retrieval-augmented finetune a VLA with an additional action-interpolation layer (like in \citep{kaustubh_regent_iclr}) to create a \RICL{}-VLA which has been primed to use its context effectively. 
%A \RICL{}-VLA 
can 
improve the base VLA's performance for any target tasks without a
%tackle entirely new tasks — featuring unseen objects, novel motions, and new scenes — without needing a 
single gradient update, instead adapting
%. Instead, it adapts 
purely through retrieval-augmentation and in-context learning, with only 10-20 demonstrations in its retrieval buffer. 
% \jd{I feel like we say RAG and ICL a few too many times in the text and figures etc. Can we just say RAG alone most of the time? Or ICL alone? If we need to, we could just remind the reader that we are actually doing RAG+ICL once in a while. Otherwise,  it's just awkward to keep saying RAG+ICL.} % \ks{either doesnt make sense wihtou the other, but if I had to really choose, RAG alone maybe makes sense}
% \jd{how many new tele-operated demonstrations per new task?} %\ks{20 only}
%, unlocking few-shot generalization. 
%, though still with limited reliability
%\RICL{} can also improve the success rate of a VLA on the long tail of training tasks.
We demonstrate this on
%in-context learning for
%the same on 
various manipulation tasks depicted in Figure \ref{fig:qualitative_results} where a state-of-the-art (SOTA) VLA fails but our \RICL{}-VLA adapts simply via RAG and ICL. 
% Next, in order to push this line of thinking further we ask the following question:  
% What happens if we perform further \RICL{}-like finetuning of the \RICL{}-VLA on task-specific demonstrations?
We further find that it is possible to get even better task performance by ``finetuning like you pretrain''~\citep{finetune_like_you_pretrain}: we optimize the \RICL{} objective on the same demonstrations as used above for ICL, and get large performance boosts.
\vspace{-0.5em}
\section{Related Work}
\label{sec:regentic_tuning_related}
\vspace{-0.5em}

\textbf{Training VLAs and multi-task generalist agents: }
%Recent work in the robot learning community has been aimed at 
There has been a spate of work in recent years on training multi-task agents in simulated settings like games~\citep{gato, kaustubh_regent_iclr, JAT, mtt, genie, icrt} and in recent months on
training VLAs for robotics~\citep{pi0_fast, pi0_diffusion, OpenVLA, Octo, RT2, RTX, robocat, gemini_robotics}. To our knowledge, there are only three prior attempts to train general agents with in-context learning (ICL) capabilities~\cite{mtt, kaustubh_regent_iclr, icrt}, none for general-purpose robotics. This is the focus of \RICL{}: we show how to post-train 
% \jd{? post-train? mid-train?} 
a pre-trained VLA to effectively learn in-context.

%Similarly, recent work in the broader reinforcement learning community has focused on training multi-task generalist agents \citep{gato, kaustubh_regent_iclr, JAT, mtt, genie}. 
%While all of these works include the capability to be finetuned to novel settings, only a couple \citep{kaustubh_regent_iclr, mtt} are capable of in-context adaptation.

\textbf{In-context learning for robotics: } 
The in-context learning abilities of large language models (LLMs) and vision-language models (VLMs) have already been found to be very useful in robotics:  
%ICL has been leveraged in the robotics community both by training policies to learn in-context and by leveraging off-the-shelf LLMs and Vision-Language Models (VLMs). 
%In the former category, we find early work in LLMs that retrains LLMs to become better in-context learners \citep{metaicl, PICL}. Towards robotics, REGENT \citep{kaustubh_regent_iclr} and MTT \citep{mtt} train general embodied agents to play held-out games. We discuss these in more detail in Section \ref{sec:regentic_tuning_background}.
%In the latter category, the robot learning community has explored agents that leverage the ICL capabilities of off-the-shelf Large Language models (LLMs) 
with suitable representations (such as keypoints or code), 
these models can in-context learn imitation policies~\citep{kat, r+x, roboprompt, llm_pattern_machines} or value functions~\citep{Eureka, GVL}. 
%to perform imitation \citep{kat, r+x, roboprompt, llm_pattern_machines} or generate value functions \citep{Eureka, GVL}. 
But, using LLMs and VLMs
% \jd{did you mean VLMs?} 
requires these methods to run completely (or mostly) open-loop \citep{kat, r+x, roboprompt} and using only LLMs makes them lose significant visual information \citep{kat, roboprompt}. Both these drawbacks affect their ability to adapt. 

\vspace{-0.5em}
\section{Background on ICL, RAG, and \texorpdfstring{$\pi_0$}{Pi0} VLAs} 
\label{sec:regentic_tuning_background}
% \jd{Should there be a background section on what in-context learning means, and how it might work for robots in particular, and why retrieval-augmentation is a good fit for this beause otherwise the context would have to hold all of the demo trajectories? Also mention prior attempts at in-context learning in robotics, e.g. applied straight to VLMs, Jason's GVL for value predictions, Eureka series for reward functions in code, Jasons'}
\vspace{-0.5em}

\textbf{In-Context Learning (ICL) }is the property of sequence to sequence models that allows them to predict an output (such as an action $\hat{a}_t$) for a new input (such as state $s_t$) given a few examples of input-output pairs in the context (such as state-action pairs $\{(s^{\prime}, a^{\prime}), (s^{\prime\prime}, a^{\prime\prime}), ...\}$). The input to the model is the concatenation of the context and the new input. %\jd{you probably should say something about what it means to ``give'' a few examples. i.e. the input to the model is  the concatenation of the context and the ... }

\textbf{Retrieval-Augmented Generation (RAG) }refers to the strategy commonly used to help LLMs predict an answer for a query. This is achieved by obtaining the information necessary for the answer by searching through a dataset and then placing said information in the context of an LLM. %\jd{split into two sentences.}
%\jd{maybe it's appropriate to say how RAG may be used in tandem with ICL? i.e. RAG to fetch input-output examples? Or are we not treating that as background? (The intro etc. treat that as background, vs our contribution.)}

Of the three methods for embodied agents mentioned in Section \ref{sec:regentic_tuning_related} that learn to in-context learn, MTT \citep{mtt} and ICRT \citep{icrt} place a few complete demonstrations in their context while REGENT \citep{kaustubh_regent_iclr} retrieves select states and actions (from the same few complete demonstrations) to place in its context. The RAG+ICL method employed by REGENT outperforms the former ICL method across held-out games and held-out simulated robotics tasks. 
REGENT demonstrates that combining the two i.e. retrieving specific examples to place into context from a demonstration dataset offers a computationally less intensive, higher performing alternative to directly placing demonstrations in context. We adopt this idea in \RICL{}.

\textbf{$\pi_0$-FAST }\citep{pi0_fast} is a state-of-the-art auto regressive VLA model that takes images, language instruction, and proprioceptive state as input and predicts actions. It can be deployed (in a variety of scenes) zero-shot on robot embodiments that are a part of its training data and few-shot after finetuning on new robot embodiments. \textbf{$\pi_0$-FAST-DROID }\citep{pi0_fast} is a VLA that was created by further finetuning $\pi_0$-FAST on the large DROID dataset \citep{droid}. The DROID dataset was collected with the Franka DROID platform, shown in Figure \ref{fig:franka_droid_setup}, across many research labs (additional details in Appendix \ref{app:more_background_regentic_tuning}).

\section{\RICL{} and creating in-context learning capable \RICL{}-VLAs}

\begin{figure}[t!]
    \centering
    \includegraphics[width=\linewidth]{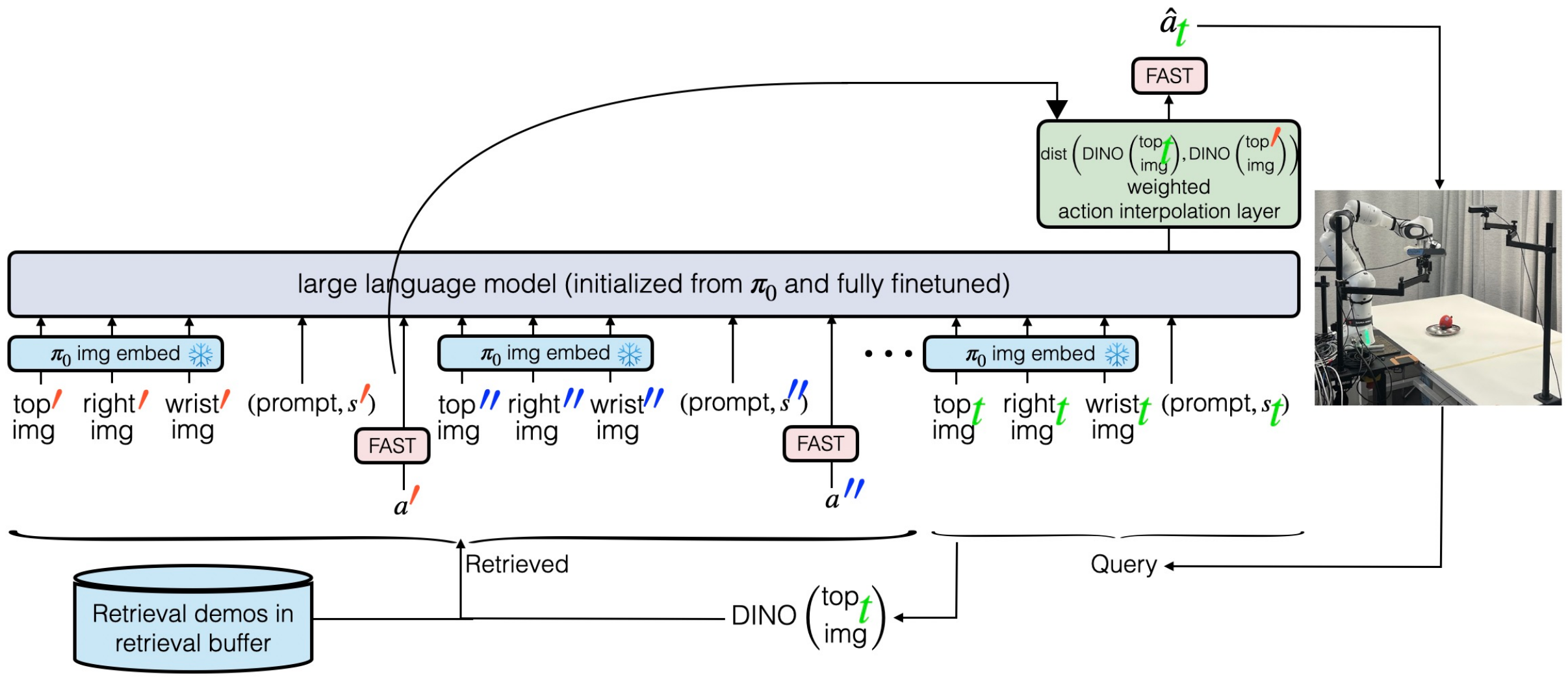}
    \caption{\small Architecture of \RICL{}-VLAs, specifically that of \RICL{}-$\pi_0$-FAST. 
    % \jdc{This figure seems to waste lots of white space, particularly vertically. Any way to improve?} 
    }
    \label{fig:regentic-pi0-arch}
    \vspace{-1em}
\end{figure}

This work aims to combine the best of both worlds from Section \ref{sec:regentic_tuning_background}--\textit{i.e.}, it aims to quickly convert a VLA that can be generally deployed 
% \jd{what does generally deployed mean? Presumably with the ``best of both worlds'' narrative, you're saying that REGENT cannot be ``generally deployed''?} \ks{yes, I'm saying REGENT is not a foundation model in the scale of pi0}
like $\pi_0$-FAST-DROID into one that also has in-context adaptation capabilities like REGENT.
Once an in-context learning capable VLA has been created, ``teaching'' it to improve its performance on a new task is as simple as downloading the model, collecting a few demonstrations, and providing them as a retrieval dataset. Then, the in-context learning capable VLA should instantly have much better success rates on this new task than the baseline VLA. 
% \jd{this feels like the introduction paragraph to the method section before the first subsection begins. Not sure it has the legs to be a section in its own right.}

\textbf{Re-training for In-Context Learning (\RICL{}): } \RICL{} enables the aforementioned conversion of a pre-trained VLA to a in-context learning capable-VLA (that we call a \RICL{}-VLA). In \RICL{}, a VLA is post-trained 
% \jd{mid-tuned? post-tuned? let's be consistent if using any of those terms.} 
on sequences of query images/states and many images, states, actions, and action obtained from the retrieval demonstrations %\jd{not sure what this means. I don't think it's clear here that you're referring to neighbors retrieved from the retrieval dataset of the previous (problem setup) section. Some notation would probably help.} 
as depicted in Figure \ref{fig:regentic-pi0-arch}. The query information at time $t$ consists of three images (top image$_t$, side image$_t$, wrist image$_t$), a language prompt describing the task, and proprioceptive state $s_t$. We use the term "query" following terminology from RAG for LLMs. The retrieved neighbors also consist of three images, the same text prompt, proprioceptive state, and action chunk (\textit{i.e.} an array of actions over many time steps).
% \jd{what is an action chunk? don't assume this is unambiguous.} 
The retrieved information is placed in the context with the closest neighbor (to the query) on the left and farther away neighbors towards the right. The closest neighbor's images, states, and actions is represented with a single $\prime$, the second closest with a double $\prime\prime$ and so on (see Figure \ref{fig:regentic-pi0-arch}). This finetuning utilizes a few "priming" demonstrations. These demonstrations are called "priming" demonstrations since their role is to prime 
% \jd{What does ``priming'' mean exactly and how is it different from ``training'' / ``finetuning''? Priming in psychology means something different from what you're saying, if that's the connection you're trying to draw.} \ks{I simply use this to refer to the training demos used for teaching the VLA to use its context but cant use the word "teaching" because the retrieval demos seem more like "teaching" demos for a target task.}
the VLA to use its context effectively. 
% \jd{I don't understand this framing. Could we first emphasize the method of converting VLA to accept retrieved context as part of its input, and then we can talk about the training data? The thing about the DROID dataset and how it cannot be easily used to create the priming data because it is not grouped into tasks feels like an implementation detail. It can either come towards the end of this section, or at the beginning of experiments.}
Further, as depicted in Figure \ref{fig:regentic-pi0-arch}, only the LLM is finetuned during \RICL{} while the image encoder is kept frozen. 
% \jd{I think the standard visual motif for this is to put a snowflake on the img embed blocks to indicate that they are ``frozen''.}
\RICL{}-VLAs perform retrieval by embedding only the top query image with an off-the-shelf DINO-v2 \citep{dino-v2} image encoder and comparing it with the embeddings of top images of the demos in the retrieval buffer with an $\ell_2$ distance metric. %\jd{first, this feels like a report: ``here's what we did'', rather than: ``here's why this is the right way to do it''. second, I might move this to implementation details anyway.} \ks{can move in camera ready; unsure where now}

Like \Regent{} \citep{kaustubh_regent_iclr}, the predicted 
% query\jd{why do you call it a ``query''} 
action $\hat{a}_t$ involves a distance-weighted interpolation between the action tokens of the closest retrieved action $a\prime$ and the final output of the large language models. We refer to this as the action interpolation layer 
% \jd{nit, but the green box is not called this anywhere in the figure.} 
and depict it within the green box above the LLM in Figure \ref{fig:regentic-pi0-arch}. This distance corresponds to the distance between the DINO embeddings of the query top image and the closest retrieved top image. The action interpolation layer assumes a maximum number of action tokens numbering $N_{\text{act}}$ and combines the one-hot encoding of each token of $a^\prime$ with the corresponding token output by the LLM $\pi_{\theta}(\text{retrieved, query})$ as follows:
\begin{align}
    \pi^{\theta}_{\text{\RICL{}-VLA}}(\text{retrieved, query}) = e^{-\lambda d} \; \text{one-hot}(a^\prime) + (1 - e^{-\lambda d}) \sigma\left(\pi_{\theta}(\text{retrieved, query})\right)
\end{align}
where $\sigma$ represents the Softmax function and $d$ denotes the $\ell_2$ distance between the DINO embeddings of top image$_t$ and its nearest neighbor top image$^\prime$. The \RICL{}-VLA performs the above interpolation for each of the $N_{\text{act}}$ tokens. These tokens are then detokenized by the FAST tokenizer to obtain an action chunk that can be executed on the robot.

Unlike the process to train \Regent{} \citep{kaustubh_regent_iclr}, \RICL{} on the other hand, only predicts and minimizes the cross-entropy loss over the query (prompt, $s_t$) tuple and predicted action chunk during training whereas \Regent{} \citep{kaustubh_regent_iclr} predicted and minimized the loss over all retrieved and query actions. 

% Also, \RICL{}-VLAs also do not have to deal with different distance values to normalize the embedding distance used in the action interpolation layer and use a single value (the maximum distance in all \RICL{} input sequences) throughout \RICL{} and deployment. \jd{this is too in-the-weeds and lacks context to make sense of it. }

The \RICL{}-VLA, after having been primed to use its context in \RICL{}, can now be deployed on a target task, which can include unseen objects and novel motions, 
% \jd{I've been using the framing of ``can be improved on a target task'' as the stock statement, and sometimes adding ``, including those involving unseen blah-blah''. mainly because I think the notion of unseen etc is a bit hard to pin down for a ``foundation model''}
with just a few task-specific demonstrations for RAG and ICL, and without any further training on those demonstrations.

\textbf{Further finetuning of a \RICL{}-VLA: }If a \RICL{}-VLA is further finetuned on those target task demonstration--that it was only retrieving from and throwing into its context previously--it can significantly improve its performance, outperforming a VLA directly vanilla finetuned on those unseen task demonstrations. This finetuning process on the few task-specific demonstrations is done exactly like \RICL{}-- \textit{i.e.}, a retrieval-augmented finetune of the \RICL{}-VLA (which has the action interpolation layer) with the same objective of minimizing the cross-entropy loss over the query (prompt, $s_t$) tuple and predicted action chunk. At deployment, the finetuned \RICL{}-VLA still retrieves from the same data that it is finetuned with, \textit{i.e.} no extra data (hyperparameters in Appendix \ref{app:more_details_regentic_tuning}). 
% \jd{goes by quickly. I'd say a bit more, or add a paragraph header or something if you want people to not miss this. Also make sure to mention that the eventual model still retrieves from the same demos that it is finetuned with, i.e., no extra data.}

\vspace{-0.75em}
\section{Experimental Setup}
\label{sec:expt_setup_regentic_tuning}
\vspace{-0.75em}

% \jd{The blacktriangleright stuff seems a bit unusual. I've used them in the past, but it feels like you have more straightforward organization schemes available here that are missing in the draft. Like an ``Experimental Setup'' section that has paragraph headers for Datasets, Tasks (including randomization and demonstration data description), Baselines, Performance Metrics}

\begin{wrapfigure}{l}{0.5\textwidth}
    \vspace{-1em}
    \includegraphics[width=0.51\linewidth]{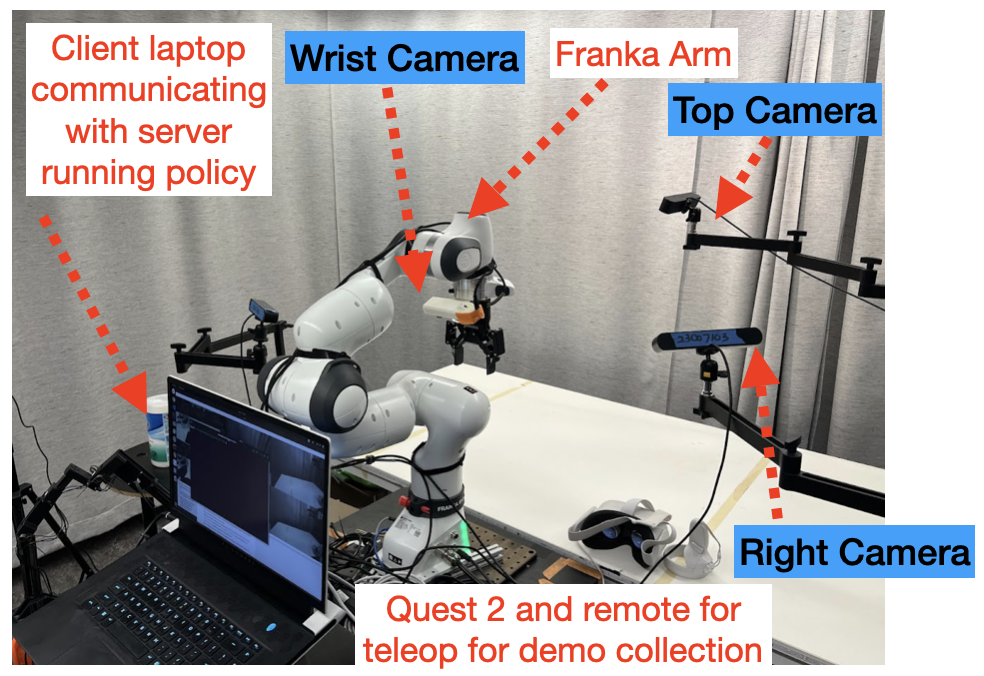} \hfill 
    \includegraphics[width=0.45\linewidth]{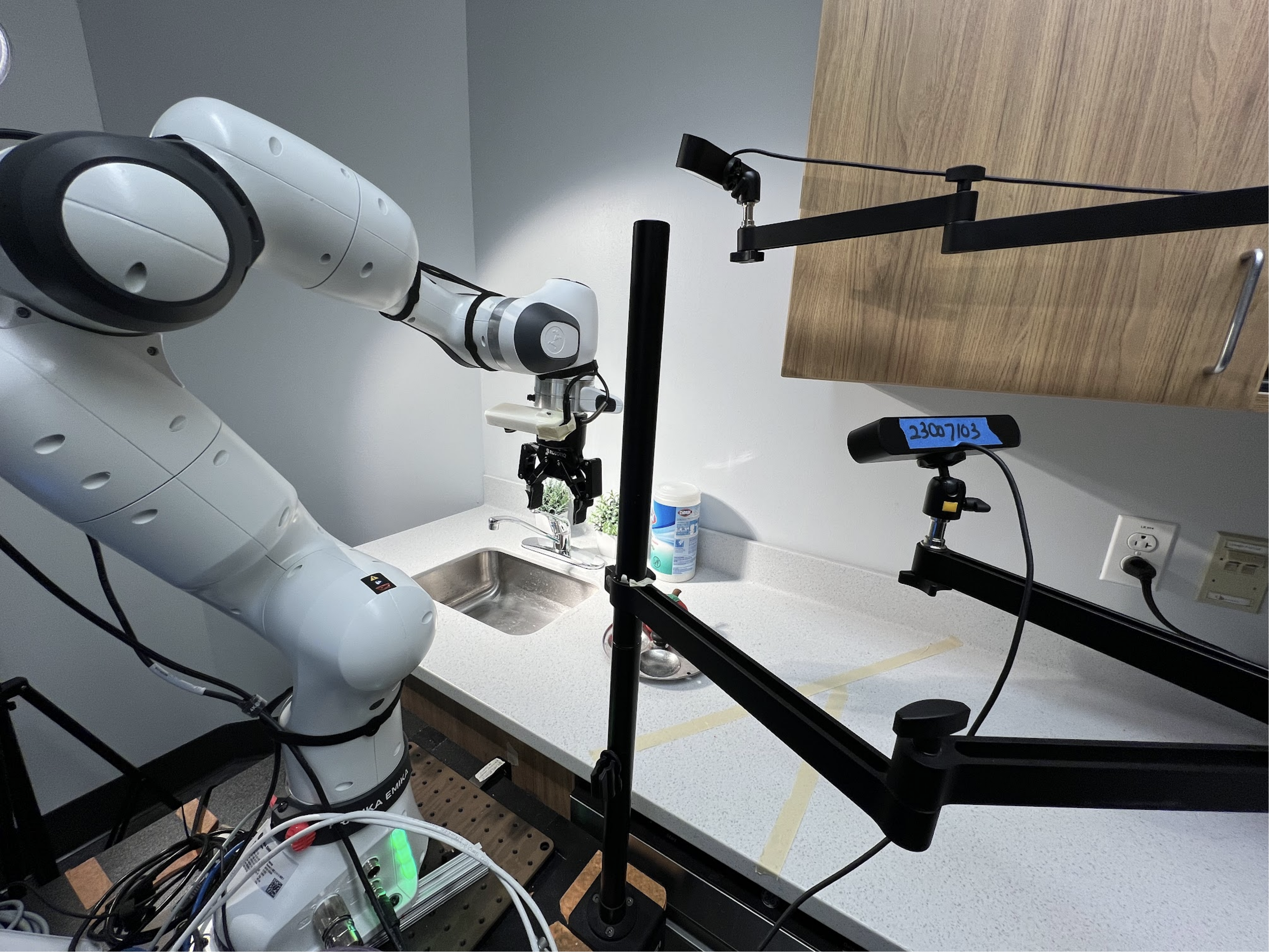}
    \caption{\small [LEFT] Our Franka DROID setup, annotated. [RIGHT] Franka DROID, including the top camera and right camera, moved to a new scene (kitchen sink).}
    \label{fig:franka_droid_setup}
    \vspace{-1em}
\end{wrapfigure}

\textbf{Training (\textit{a.k.a.}, Priming) data for \RICL{}: } 
% \jd{reference the figure showing the setup here. also, I think (not sure) it might be more appropriate to move the figure closer to here, even though you briefly reference it earlier. }
% \RICL{} requires priming demonstrations in a few tasks and specifically requires atleast a few demonstrations in each training task. This is because, during \RICL{}, each state in each demonstration from each training task is treated as the query state and all the other demonstrations from that same task are used as retrieval demonstrations. We simply train in the same way we deploy the model. Since such a training dataset is not readily available, we collect our own dataset with the Franka DROID platform. 
We collect 20 demonstrations with the Franka DROID platform (see Figure \ref{fig:franka_droid_setup}), 
% \jd{``the tabletop''? Isn't this the first time you're mentioning this? } \ks{removed}
% \jd{This is probably where I would first bring up that, ideally, you would just use any task-wise pre-collected robot training data lying around, but given that there wasn't a convenient dataset, you collected your own.} 
randomizing the initial position of the primary object, in 20 pick and place tasks (total 400 demonstrations). The exact list of tasks, an image of all the objects used, and more details about the platform are in Appendix \ref{app:more_details_regentic_tuning}.  We perform \RICL{}-- starting from the weights of $\pi_0$-FAST-DROID, fully finetuning its LLM, and keeping its image encoder frozen-- with the hyperparameters detailed in Appendix \ref{app:more_details_regentic_tuning}. We call the model obtained after three epochs of training as \RICL{}-$\pi_0$-FAST-DROID.

In each task, for each demonstration, and for each state in that demonstration, we use that state as the query and retrieve four neighbors from the other 19 demonstrations to create training input sequences. In this way, we end up training the model in the same way we would deploy the model. We collect data as detailed above since such a dataset is not available.

\textbf{Evaluation tasks with unseen objects, novel motions, different scenes: }We evaluate all methods on the following tasks, which 
involve
%have 
a task-relevant unseen object  and/or a completely novel motion in two different scenes (tabletop and kitchen sink). ``Unseen'' here means that these objects and motions are not in either the \RICL{} priming data or the DROID data~\citep{droid}.
%We note that all unseen objects and novel motions below are not in the above \RICL{} priming data or the DROID dataset \citep{droid}. 
We checked the latter by searching over the DROID dataset's language annotations (and some recordings when language was not adequate). First, we start with a simple task that primarily tests language grounding.
\begin{itemize}[leftmargin=*]
    \item (\texttt{pokeball}) "pick up the pokeball and put it in the tray": has an unseen object (pokeball) with a couple of distractors on the table. 
    % \jd{maybe come up with simple names for each like \texttt{poke}, for easy future reference?}
\end{itemize}
Next, we test on simple tasks that test both language grounding and adaptation to novel motions.
\begin{itemize}[leftmargin=*]
    \item (\texttt{idliplate}) "move the idli plate to the left": has an unseen object (idli plate) with a apple (distractor) sitting on the plate. It also requires the robot to do an unfamiliar grasp to move the uniquely shaped plate with depressions to the right.
\end{itemize}
\begin{itemize}[leftmargin=*]
    \item (\texttt{squeegee}) "move the squeegee to the right and try to drag it": 
    % \jd{this does not seem to specify that you want to drag it in the way that would actually clean a surface -- you could drag it in other ways, right?} \ks{the other squeegee task takes care of that where it is clearly told to clean, I think its fine}
    has an unseen object (squeegee). It also requires the robot to do a novel motion of slightly lifting the handle of the squeegee while keeping its rubber on the table to drag it across the table. 
\end{itemize}
We then test on versions of the above two simple tasks in a new scene (kitchen sink area) with new camera positions/orientations (no calibration necessary), new lighting, and new distractors.
\begin{itemize}[leftmargin=*]
    \item (\texttt{sink-idliplate}) "move the idli plate to the sink": all of the previous challenges \& a new scene.
    \item (\texttt{sink-squeegee}) "use the squeegee to clean the counter": all of the challenges of the previous task with pellets to be cleaned up on the counter in the new scene.
\end{itemize}
We also test on the long-tail of the training task distribution. These tasks consist of variations of objects in the DROID dataset. These object classes appear infrequently in the dataset.
\begin{itemize}[leftmargin=*]
    \item (\texttt{toaster}) "push the lever on the toaster": has a different version of an object (a particular toaster brand) in the DROID dataset. This task tests the long tail of the training task distribution. It also requires a precise placement and movement to push the lever down.
    \item (\texttt{door}) "open the door of the bottom shelf":
    % \jd{my immediate response to this task description is that it probably is also a grounding problem since phrases like ``bottom shelf'' are not easily resolved by most VLMs today.} 
    has a different version of an object (a particular shelf) in the DROID dataset. This task also is in the long tail. It requires a novel and precise motion that can handle the large top shelf acting as an obstacle when reaching the bottom door's handle.
\end{itemize}
Finally, we test on a longer horizon task that is a composition of many simple tasks.
\begin{itemize}[leftmargin=*]
    \item (\texttt{bagel}) "pick up the bagel and put it in the toaster": has an unseen object (bagel). It also requires the robot to do a composite novel motion--an unfamiliar grasp on the edge of the bagel, the twist-and-lift motion, and placing in the slot. 
    % We note the height since none of the \RICL{} priming data has receptacles at the height of the toaster. 
\end{itemize}
% \jd{One weakness of the paper is that it's a bit underwhelming to evaluate a ``generalist'' model on some 6 very specific tasks. A way to overcome this is to motivate why this is a good collection of tasks. What are they looking to evaluate? Are they just random tasks you thought of? Is there a systematic thought process in mind when picking this collection of tasks? Does the baseline policy pi-zero perform poorly at all of these tasks? What happens if you pick a task for which the baseline policy is already good? How do these tasks differ from those for which pi-zero has been demonstrated to work best?}
% \ks{organize the tasks based on language grounding issue (new object only) / adaptation issue (new objkect + motion) / (new objec + new motion + new scene) / can organize based on diffculty like compound task [bagel into toaster is like a two step] can help understand figure 5}

\textbf{Retrieval data in evaluation tasks: }In all the evaluation tasks, we collect 20 demonstrations, randomizing the initial position of the primary object, for \RICL{}-$\pi_0$-FAST-DROID to retrieve from and throw in its context to adapt to the task.

\textbf{Evaluation metrics and comprehensive randomization: } We collect 10 test rollouts for all methods on all evaluation tasks with randomly chosen initial positions and orientations in each rollout. We set these intial positions and orientations all across the table. The distractors, if any, are kept approximately in the same region of the table but they are also not fixed in place. We calculate the success of the full tasks, in addition to tracking intermediate checkpoints for a better understanding of the progress of each method on each task. 
% \jd{do you fix the same set of 10 initial conditions for all compared approaches on each task? If not, I'd suggest doing that after submission. Otherwise, 10 different random trials for each method is going to likely be too high-variance for any reliable comparisons.} \ks{they are approx same across tasks}

% Our results already include various measures of robustness \jd{Not sure what you mean. If you just want to say that the experiments invlved variations in these things, then this belongs in the task description not here. If that's not the case, then what exactly are the measures you;re using to quantify robustness measures?} such as to (1) various initial positions and orientations as discussed above, (2) two different environments: tabletop and kitchen sink (3) various lighting conditions including differences in lighting when collecting retrieval demos and performing test rollouts, (4) various table surface conditions and background curtain positions, and (5) changes in camera position both when moving to the sink environment and over over time. We also observe robustness to a human hand and other random tools (like a pole, a stick, a reacher grabber, badminton and squash rackets, etc) that were stuck into the scene to move objects on the table in our qualitative  demonstrations seen in Figure \ref{fig:qualitative_reliability_results}.

\textbf{Baselines and ablations: }We compare \RICL{}-$\pi_0$-FAST-DROID with vanilla $\pi_0$-FAST-DROID on all tasks. We also compare with 'Retrieve and play', a 1 nearest neighbor baseline from \citep{kaustubh_regent_iclr}, which simply outputs the first retrieved action $a^\prime$. We also compare with a trained-from-scratch Diffusion Policy baseline. We perform these two comparisons on the simpler evaluation tasks (\texttt{pokeball}, \texttt{idliplate}, \texttt{squeegee}). Upon observing their low success rates, we leave them out for the more complex tasks. In the tasks performed in a new scene (\texttt{sink-idliplate}, \texttt{sink-squeegee}), we  aim to test \RICL{}'s ability to retain $\pi_0$'s helpful scene-generalization capabilities while adapting in-context to a new task and hence only test these two methods.
We also ablate the number of demonstrations used by each method in the \texttt{idliplate} task.
% \jd{baselines and ablations?}

\textbf{Further finetuning: }We further finetune \RICL{}-$\pi_0$-FAST-DROID on each evaluation task on the 20 demonstrations collected for retrieval. For comparison, we also further (vanilla) finetune $\pi_0$-FAST-DROID on these same 20 demonstrations in each task. 
% This means that many evaluation tasks have their own  finetuned checkpoints of both \RICL{}-$\pi_0$-FAST-DROID and $\pi_0$-FAST-DROID.

% \textbf{Checking for loss of capabilities: }Finally, we check if \RICL{}-$\pi_0$-FAST-DROID can be deployed without any task-specific retrieval demonstrations on three simple tasks that $\pi_0$-FAST-DROID can perform out-of-the-box. These tasks are "move the can to the tray", "pick up the marker and put it in the tray", and "place the apple next to the can".

\section{Experimental Evaluation}
\label{sec:experiments_regentic_tuning}

\begin{figure}[t!]
    \centering
    \includegraphics[height=4cm]{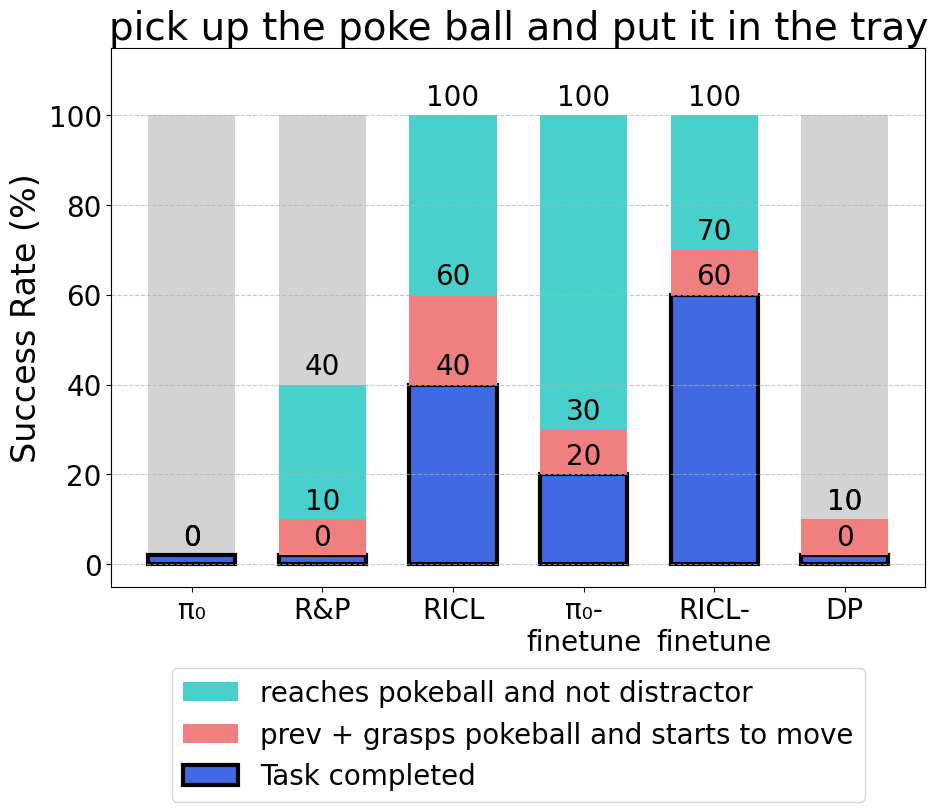}
    \includegraphics[height=4cm]{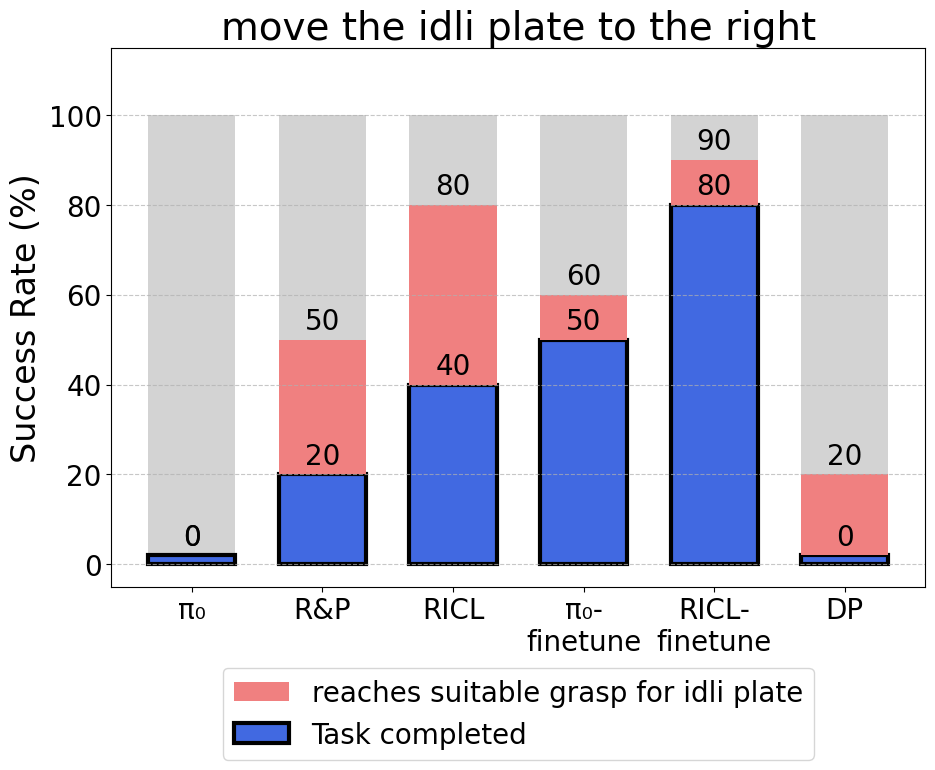}
    \includegraphics[height=4cm]{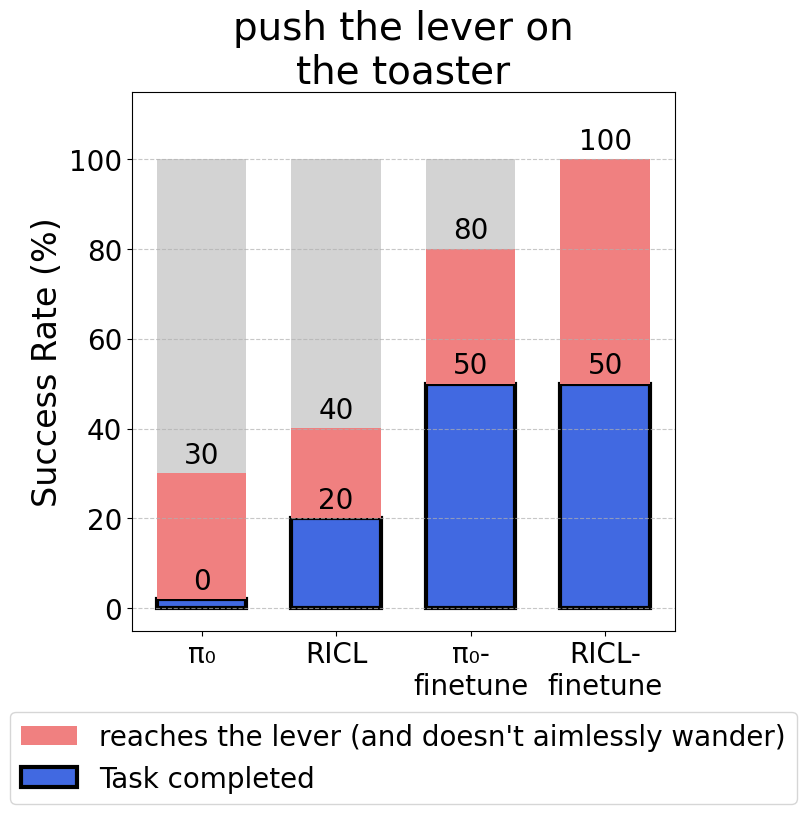}
    \includegraphics[height=4cm]{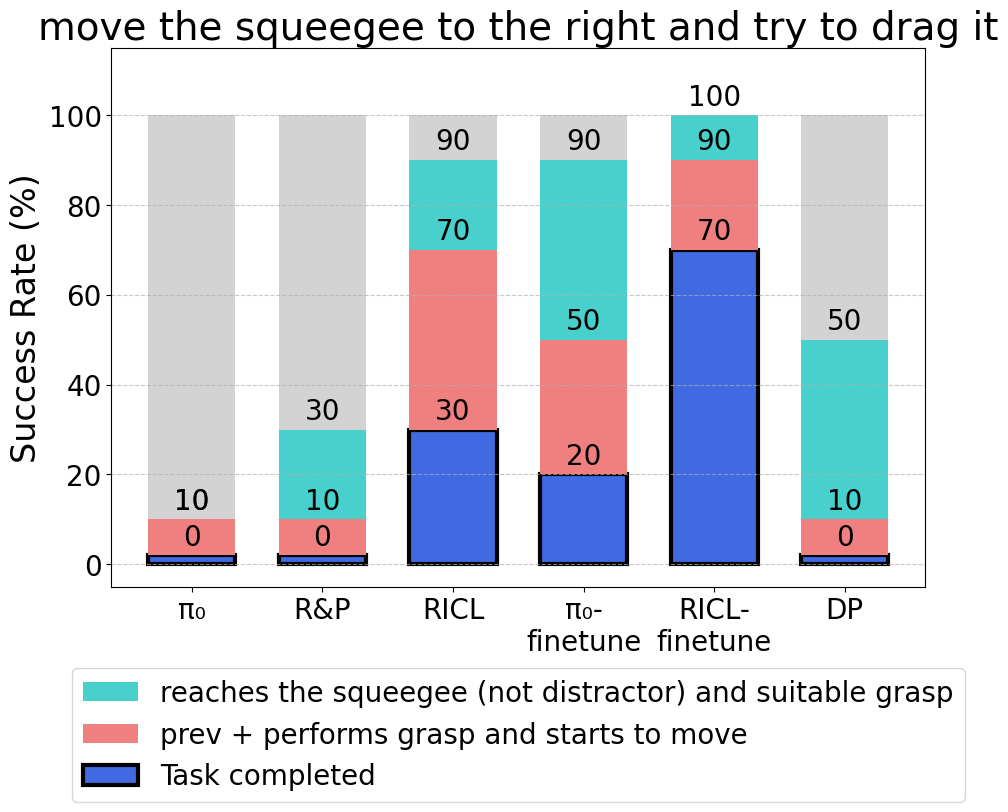}
    \includegraphics[height=4cm]{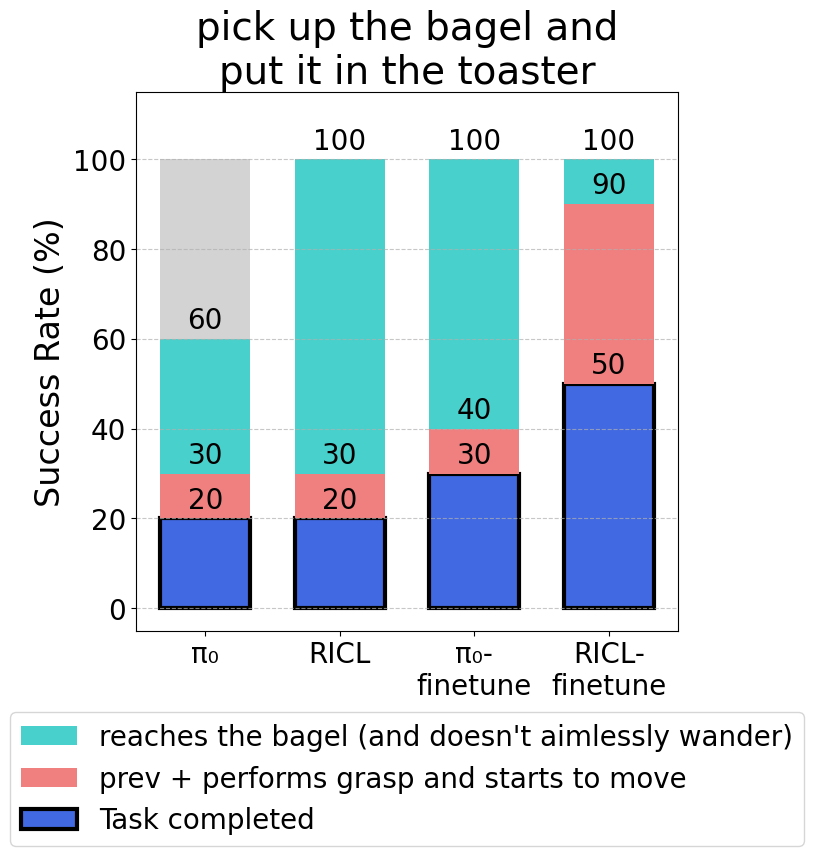}
    \includegraphics[height=4cm]{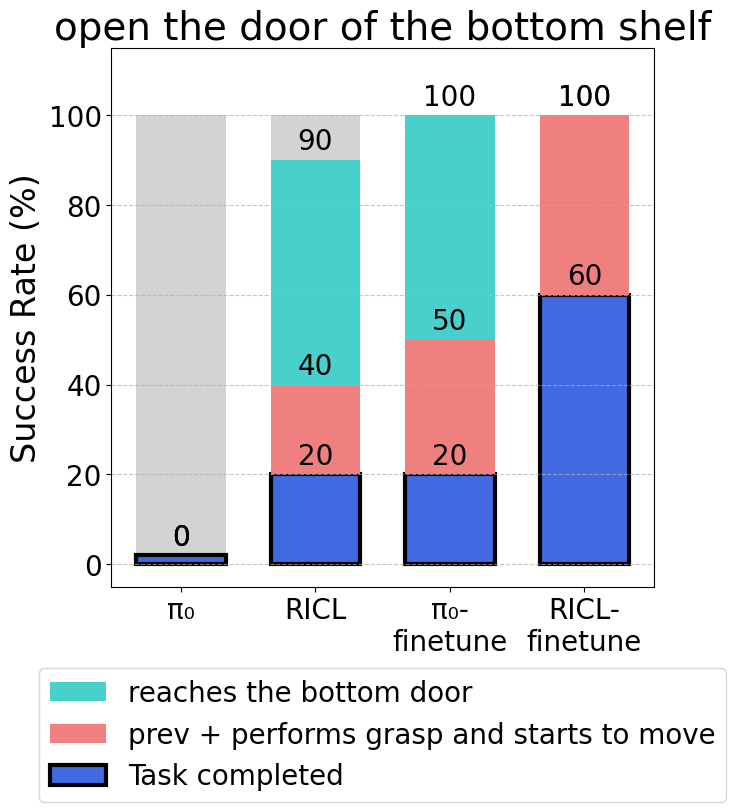}
    \includegraphics[height=4cm]{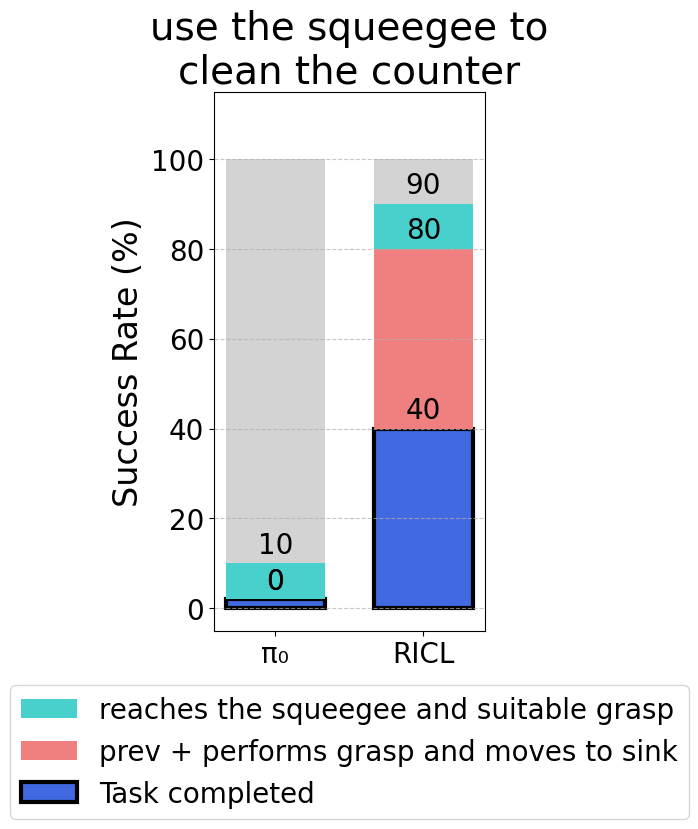}
    \includegraphics[height=4cm]{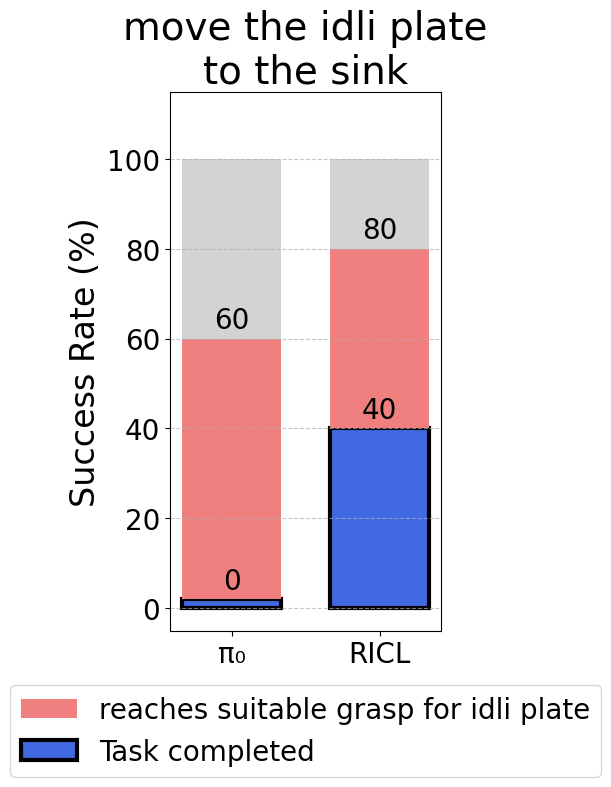}
    \includegraphics[height=3.5cm]{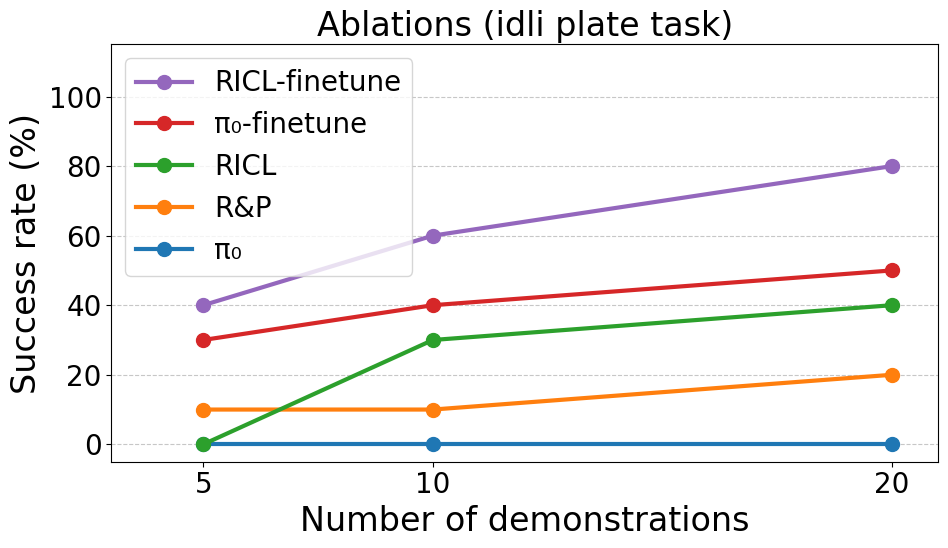}
    \caption{\small Success rates of 10 test rollouts from various methods across various tasks represented by stacked bar plots. The lowest bar (dark blue) in each stacked column represents full task success rate, and other bars are the success rates for reaching earlier waypoints. Gray regions represent the fraction of runs that did not even reach the first waypoint for the task. %represents full task completion and is most important.
    %The upper bars help indicate the progress of each methods' rollouts towards completing the task. 
    We note that $\pi_0$ refers to $\pi_0$-FAST-DROID and RICL to \RICL{}-$\pi_0$-FAST-DROID in the plots. We also plot the performance of various methods vs the number of demonstration in the \texttt{idliplate} task on the bottom right.
    }
    \label{fig:quantitative_results}
    \vspace{-1.5em}
\end{figure}
\begin{figure}[t]
    \centering
    \begin{subfigure}{\textwidth}
        \centering
        \includegraphics[width=0.75\linewidth]{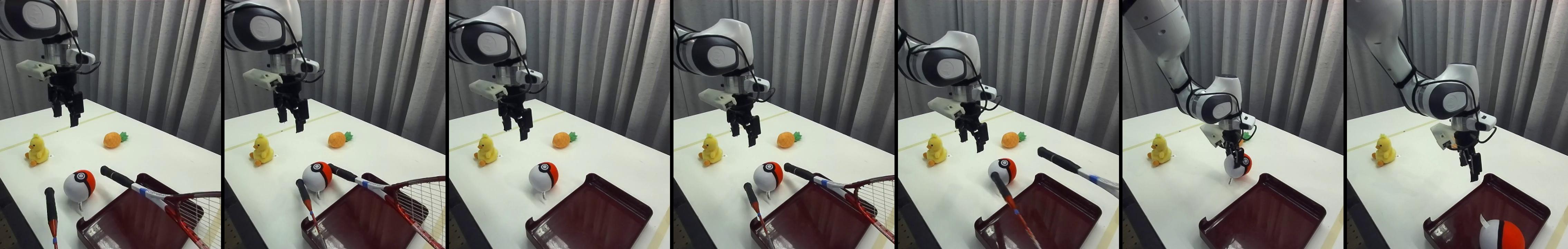}
        \vspace{-0.5em}
        \caption{\scriptsize Task: "pick up the poke ball and put it in the tray"}
    \end{subfigure}
    \begin{subfigure}{\textwidth}
        \centering
        \includegraphics[width=0.75\linewidth]{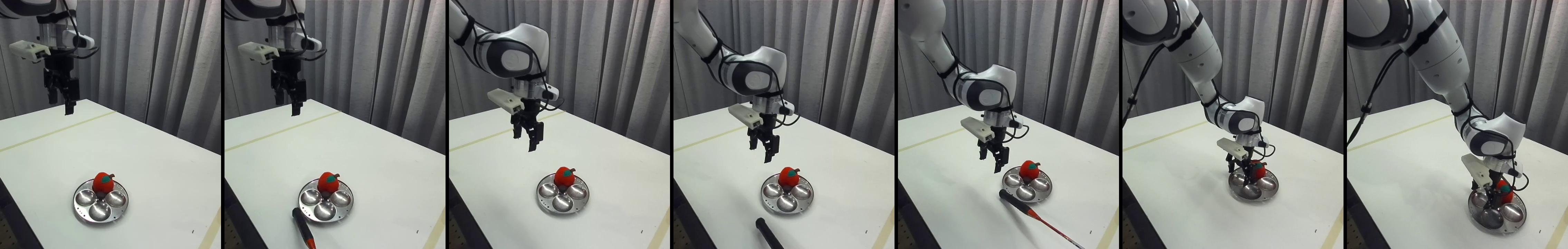}
        \vspace{-0.5em}
        \caption{\scriptsize Task: "move the idli plate to the right"}
    \end{subfigure}
    \begin{subfigure}{\textwidth}
        \centering
        \includegraphics[width=0.75\linewidth]{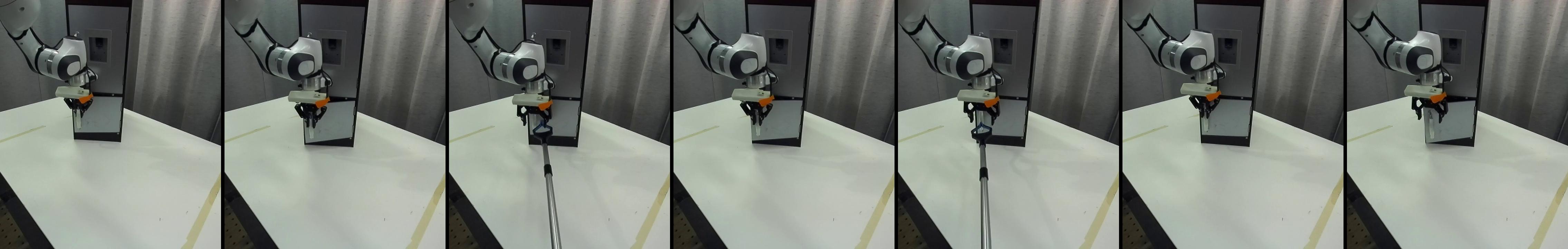}
        \vspace{-0.5em}
        \caption{\scriptsize Task: "open the door of the bottom shelf"}
        \vspace{-0.5em}
    \end{subfigure}
    \caption{\small Qualitative visualization of the reactivity and robustness of \RICL{}-$\pi_0$-FAST-DROID-finetuned on 20 task-specific demonstrations in a dynamic test rollout. In the above, a human randomly perturbs and displaces the primary object during the test rollout. Additional results can be found in Figure \ref{fig:qualitative_reliability_results_appendix} in Appendix \ref{app:additional_results_regentic_tuning}.}
    \label{fig:qualitative_reliability_results}
    \vspace{-1em}
\end{figure}

% \begin{figure}[t!]
%     \centering
%     \includegraphics[height=3.5cm]{regentic_tuning/plots_dont_delete/Ablations_idli_plate_task.png}
%     % \includegraphics[width=\linewidth]{regentic_tuning/plots_dont_delete/colors_2.png}
%     \caption{Success rates of 10 test rollouts from various methods in the \texttt{idliplate} task for various number of demonstrations to retrieve from (for \RICL{}-$\pi_0$-FAST-DROID and Retrieve and Play) or finetune on (for \RICL{}-$\pi_0$-FAST-DROID-finetuned or $\pi_0$-FAST-DROID-finetuned). %The two subtasks indicate the progress of each method towards completing that task given those number of demonstrations to retrieve from or finetune on.
%     }
%     \label{fig:ablations}
% \end{figure}

% \jd{There should be some clear delineation here that this is where you are beginning to talk about experimental results rather than setup. Otherwise a bit lost in the text.}
\textbf{Generalization to unseen objects and novel motions and new scenes: }We plot the quantitative results across tasks and methods in Figure \ref{fig:quantitative_results}. We observe that \RICL{}-$\pi_0$-FAST-DROID outperforms $\pi_0$-FAST-DROID, especially in earlier checkpoints of the task, but also in overall task success. 
In aggregate, across all evaluated tasks, $\pi_0$-FAST-DROID obtains a complete task success rate of 2.5\% and a checkpoint completion rate of up to 21.25\%. On the other hand, \RICL{}-$\pi_0$-FAST-DROID obtains a significantly improved complete task success rate of 31.25\% and a checkpoint completion rate of up to 83.75\%.
% \jd{could you report an aggregate success rate across all tasks here in the text for all methods?} 

We particularly note that \RICL{}-$\pi_0$-FAST-DROID has significantly improved language grounding to move towards the unseen objects just based on contextual information. More importantly, \RICL{}-$\pi_0$-FAST-DROID also overcomes the adaptation issue faced by $\pi_0$-FAST-DROID. Where $\pi_0$-FAST-DROID struggles with grasps and motions, \RICL{}-$\pi_0$-FAST-DROID demonstrates the ability to infer novel grasps and motions from its context as evidenced in six tasks (both \texttt{squeegee} tasks, \texttt{sink-idliplate}, \texttt{bagel}, \texttt{toaster}, and \texttt{door}). We plot the qualitative results depicting key test rollouts and behaviors of $\pi_0$-FAST-DROID and \RICL{}-$\pi_0$-FAST-DROID in Figure \ref{fig:qualitative_results} and Figure \ref{fig:qualitative_results_appendix}. We also provide side-by-side comparisons and detailed explanations of rollouts in the same Figures.

Unexpectedly, we observe in some tasks that \RICL{}-$\pi_0$-FAST-DROID seems to predict and execute action sequences that are not like the motions in the retrieval dataset. For example, in \texttt{idliplate}, \RICL{}-$\pi_0$-FAST-DROID moves to the left of the idli plate, closes its gripper and pushes the plate to the right. But, all 20 demonstrations in the retrieval buffer were collected with the motion of dipping the gripper into the depressions and moving the plate to the right. Hence, \RICL{}-$\pi_0$-FAST-DROID has seemingly elicited latent actions or knowledge  
% \jd{I never really understood the ``latent actions'' thing. What is latent about it? Do you mean this as shorthand for ``latent knowledge about actions'' or something like that?} \ks{yes}
% possibly from the earliest $\pi_0$-FAST pretraining, 
% \jd{Why do you say this? Are you saying how else would it know how to drag an idli plate? An alternative explanation is simply that it ``understands'' geometry well enough to ``work it out'' i.e. generalize from what it's seen. And once the language issue is resolved, it can figure out how to move?}\ks{yes, this is what I call latent actions, as in it had to faced similar geometries before to understand to do this} 
to accomplish this task (also seen in \texttt{toaster}).

\textbf{Significantly improved performance after further finetuning the \RICL{}-VLA: }We observe a significant improvement in performance after further finetuning \RICL{}-$\pi_0$-FAST-DROID on each task's 20 demonstrations. 
% In fact, \RICL{}-$\pi_0$-FAST-DROID finetuned obtains between 60-80\% complete task success rate while $\pi_0$-FAST-DROID-finetuned (on the same 20 demonstrations) only reaches a top complete task success rate of 50\%. 
In aggregate, across all evaluated tasks, $\pi_0$-FAST-DROID-finetuned obtains a complete task success rate of 31.67\%, while \RICL{}-$\pi_0$-FAST-DROID-finetuned obtains a complete task success rate of 61.67\%.
In fact, we not only see that the \RICL{}-VLA with finetuning is significantly better than the base VLA with finetuning (at almost double the aggregate performance). But we also observe comparable performance in complete task success rate, in aggregate, between the \RICL{}-VLA (at 31.25\%), which only uses RAG and ICL, and the base VLA with finetuning on the target task data (at 31.67\%).
% \jd{It's good to show that RICL + finetuning > base + finetuning. But even stronger to be able to compare plain RICL to base + finetuning, even if you can only say that they are similar in performance and we don't have an outright win. Drives home the point that even without any parameter tuning, you get (nearly?) just as good performance.}

We hypothesize that further finetuning our VLA is significantly better than doing so with the base VLA simply because our VLA can use all of its capacity to focus on interpolating amongst the retrieved images, states, and actions to predict a new action and does not have to memorize any data. This is in line with the observed parameter-efficiency \& performance advantages of RAG LLMs \citep{borgeaud2022RETRO}.
% The base VLA, on the other hand, has to use some of its capacity as a memory to remember the data it was finetuned on so that it can recollect it again at test time and also perform some kind of interpolation in the embedding space to predict a new action. \jd{I think people say similar things about RAG LLMs and why they can afford to be much smaller than plain-old LLMs. This passage would soudn less speculative if you found a reference and made this connection explicitly, like ``This is in line with the observed parameter-efficiency(?) advantages of RAG LLMs''}

We qualitatively demonstrate the reactivity and robustness of \RICL{}-$\pi_0$-FAST-DROID by randomly perturbing and displacing objects during a test rollout as shown in Figure \ref{fig:qualitative_reliability_results} and Figure \ref{fig:qualitative_reliability_results_appendix}. 

% \textbf{Task-specific discussion of performance: }

\textbf{Ablating the number of retrieval/finetuning demonstrations: }We plot the ablation results, ablating the number of demonstrations used in the retrieval buffer or for finetuning, for \texttt{idliplate} in the bottom right of Figure \ref{fig:qualitative_results}. We found that too few demonstrations (such as 5) results in \RICL{}-$\pi_0$-FAST-DROID starting to behave like $\pi_0$-FAST-DROID to the extent where in one test rollout, it too moves the apple instead of the idli plate. This demonstrates the requirement for atleast 10 demonstrations. 
% We also found a different elicitation of latent actions with \RICL{}-$\pi_0$-FAST-DROID when it has only 10 retrieval demonstrations where it grips the edge of the plate and moves it to the right. 
Also, from this figure, we conclude that more retrieval demonstrations help \RICL{}-$\pi_0$-FAST-DROID improve, towards catching up with $\pi_0$-FAST-DROID-finetuned. We also see that \RICL{}-$\pi_0$-FAST-DROID-finetuned is significantly better than $\pi_0$-FAST-DROID-finetuned at every number of demonstrations.

\textbf{No loss-of-capabilities results: } 
% \jd{This needs a little motivation. ``One might wonder: does RICL post-training come at the cost of losing the ability in the base VLA to perform without any retrieved context? To evaluate this, we do blah-blah.''}
One might wonder: does RICL post-training come at the cost of losing the ability in the base VLA to perform without any retrieval data? To evaluate this, we test \RICL{}-$\pi_0$-FAST-DROID with randomly chosen priming demonstration in the retrieval buffer, rather than any meaningful task-specific demonstrations, on three tasks: "move the can to the tray", "pick up the marker and put it in the tray", and "place the apple next to the can".
% \jd{what three tasks?} 
It obtains an 80\% success rate, just like $\pi_0$-FAST-DROID, demonstrating that \RICL{} has not led to any loss of capabilities.
\vspace{-0.5em}
\section{Conclusions and Future Work}
\vspace{-0.5em}

We have demonstrated how \RICL{} can be used to convert a VLA to a \RICL{}-VLA that can use its context to adapt to completely new tasks, including unseen objects and novel motions, with just RAG and ICL. We found the \RICL{}-VLA to even have comparable performance, in aggregate, with the base VLA finetuned on target task data. We have also demonstrated a significant boost in performance when a \RICL{}-VLA is further finetuned on task-specific demonstrations.
% , that it was previously only retrieveing from and throwing into its context, compared to a VLA vanilla finetuned on the same demonstrations.\jd{I think the comparison of RICL + finetuning to VLA + finetuning is too heavily emphasized not just here but at several places in the text, possibly missing the opportunity to instead present a more compelling story of: RICL w/o finetuning to base VLA + finetuning.} \ks{I have fixed this now}
In future work, we believe that scaling up \RICL{} in both the number of priming demonstrations and parameter size will further boost the performance of the in-context learning capable \RICL{}-VLA. 
% \jd{scale up along what axis? And what is reliability?}

\section{Limitations}

% \RICL{}-VLAs are only as powerful \jd{this seems contrary to the point we are making --- RICL can improve a VLA. Maybe you need a bit more nuance. We find that RICL does not typically enable learning any tasks that are too far beyond those that pi-zero can perform.} 
% as the underlying VLA. 
We find that \RICL{} does not typically enable learning any tasks that are too far beyond those that the base VLA can perform.
This is why in this paper we primarily focus on pick and place tasks, the primary strength of the $\pi_0$ VLAs used in this work. Further, while \RICL{}-VLAs, unlike vanilla VLAs, can adapt to new tasks in-context, they can still benefit from improved performance. We believe that scaling up \RICL{} in both the number of priming demonstrations/tasks and number of parameters can help with this issue in future work. Another limitation of \RICL{}-VLAs remains their need for a few teleoperated demonstrations. Collecting these demonstrations for every new task across settings is not scalable. We believe that videos of human demonstrations (such as in \citep{r+x}) can help bridge this gap. In our experiments, \RICL{}-$\pi_0$-FAST-DROID very easily removes the issue of language grounding for new tasks and motions but struggles with generalizing to significantly novel motions such as playing a forehand with a tennis racket. We believe future work including more diverse motions, rather than just pick and place, in the \RICL{} dataset can fix this issue. 

%===============================================================================

% \section*{Acknowledgements} 
\acknowledgments{This work was supported in part by ARO MURI W911NF-20-1-0080, NSF 2143274, NSF CAREER 2239301, NSF 2331783, and DARPA TIAMAT HR00112490421. Any opinions, findings, conclusions or recommendations expressed in this material are those of the authors and do not necessarily reflect the views the Army Research Office (ARO), the Department of Defense, or the United States Government. We also thank the anonymous reviewers for their helpful comments.}

\clearpage
% The acknowledgments are automatically included only in the final and preprint versions of the paper.
% \acknowledgments{If a paper is accepted, the final camera-ready version will (and probably should) include acknowledgments. All acknowledgments go at the end of the paper, including thanks to reviewers who gave useful comments, to colleagues who contributed to the ideas, and to funding agencies and corporate sponsors that provided financial support.}

%===============================================================================

% no \bibliographystyle is required, since the corl style is automatically used.
\bibliography{ref}  % .bib

\begin{thebibliography}{34}
\providecommand{\natexlab}[1]{#1}
\providecommand{\url}[1]{\texttt{#1}}
\expandafter\ifx\csname urlstyle\endcsname\relax
  \providecommand{\doi}[1]{doi: #1}\else
  \providecommand{\doi}{doi: \begingroup \urlstyle{rm}\Url}\fi

\bibitem[Pertsch et~al.(2025)Pertsch, Stachowicz, Ichter, Driess, Nair, Vuong, Mees, Finn, and Levine]{pi0_fast}
K.~Pertsch, K.~Stachowicz, B.~Ichter, D.~Driess, S.~Nair, Q.~Vuong, O.~Mees, C.~Finn, and S.~Levine.
\newblock Fast: Efficient action tokenization for vision-language-action models.
\newblock \emph{arXiv preprint arXiv:2501.09747}, 2025.

\bibitem[Black et~al.(2024)Black, Brown, Driess, Esmail, Equi, Finn, Fusai, Groom, Hausman, Ichter, et~al.]{pi0_diffusion}
K.~Black, N.~Brown, D.~Driess, A.~Esmail, M.~Equi, C.~Finn, N.~Fusai, L.~Groom, K.~Hausman, B.~Ichter, et~al.
\newblock $\pi_0$: A vision-language-action flow model for general robot control.
\newblock \emph{arXiv preprint arXiv:2410.24164}, 2024.

\bibitem[Kim et~al.(2024)Kim, Pertsch, Karamcheti, Xiao, Balakrishna, Nair, Rafailov, Foster, Lam, Sanketi, et~al.]{OpenVLA}
M.~J. Kim, K.~Pertsch, S.~Karamcheti, T.~Xiao, A.~Balakrishna, S.~Nair, R.~Rafailov, E.~Foster, G.~Lam, P.~Sanketi, et~al.
\newblock Openvla: An open-source vision-language-action model.
\newblock \emph{arXiv preprint arXiv:2406.09246}, 2024.

\bibitem[Team et~al.(2024)Team, Ghosh, Walke, Pertsch, Black, Mees, Dasari, Hejna, Kreiman, Xu, et~al.]{Octo}
O.~M. Team, D.~Ghosh, H.~Walke, K.~Pertsch, K.~Black, O.~Mees, S.~Dasari, J.~Hejna, T.~Kreiman, C.~Xu, et~al.
\newblock Octo: An open-source generalist robot policy.
\newblock \emph{arXiv preprint arXiv:2405.12213}, 2024.

\bibitem[Bousmalis et~al.(2023)Bousmalis, Vezzani, Rao, Devin, Lee, Bauza, Davchev, Zhou, Gupta, Raju, et~al.]{robocat}
K.~Bousmalis, G.~Vezzani, D.~Rao, C.~Devin, A.~X. Lee, M.~Bauza, T.~Davchev, Y.~Zhou, A.~Gupta, A.~Raju, et~al.
\newblock Robocat: A self-improving foundation agent for robotic manipulation.
\newblock \emph{arXiv preprint arXiv:2306.11706}, 2023.

\bibitem[Zitkovich et~al.(2023)Zitkovich, Yu, Xu, Xu, Xiao, Xia, Wu, Wohlhart, Welker, Wahid, et~al.]{RT2}
B.~Zitkovich, T.~Yu, S.~Xu, P.~Xu, T.~Xiao, F.~Xia, J.~Wu, P.~Wohlhart, S.~Welker, A.~Wahid, et~al.
\newblock Rt-2: Vision-language-action models transfer web knowledge to robotic control.
\newblock In \emph{Conference on Robot Learning}, pages 2165--2183. PMLR, 2023.

\bibitem[Padalkar et~al.(2023)Padalkar, Pooley, Jain, Bewley, Herzog, Irpan, Khazatsky, Rai, Singh, Brohan, et~al.]{RTX}
A.~Padalkar, A.~Pooley, A.~Jain, A.~Bewley, A.~Herzog, A.~Irpan, A.~Khazatsky, A.~Rai, A.~Singh, A.~Brohan, et~al.
\newblock Open x-embodiment: Robotic learning datasets and rt-x models.
\newblock \emph{arXiv preprint arXiv:2310.08864}, 2023.

\bibitem[Team et~al.(2025)Team, Abeyruwan, Ainslie, Alayrac, Arenas, Armstrong, Balakrishna, Baruch, Bauza, Blokzijl, et~al.]{gemini_robotics}
G.~R. Team, S.~Abeyruwan, J.~Ainslie, J.-B. Alayrac, M.~G. Arenas, T.~Armstrong, A.~Balakrishna, R.~Baruch, M.~Bauza, M.~Blokzijl, et~al.
\newblock Gemini robotics: Bringing ai into the physical world.
\newblock \emph{arXiv preprint arXiv:2503.20020}, 2025.

\bibitem[Brown et~al.(2020)Brown, Mann, Ryder, Subbiah, Kaplan, Dhariwal, Neelakantan, Shyam, Sastry, Askell, Agarwal, Herbert{-}Voss, Krueger, Henighan, Child, Ramesh, Ziegler, Wu, Winter, Hesse, Chen, Sigler, Litwin, Gray, Chess, Clark, Berner, McCandlish, Radford, Sutskever, and Amodei]{in-context-learning}
T.~B. Brown, B.~Mann, N.~Ryder, M.~Subbiah, J.~Kaplan, P.~Dhariwal, A.~Neelakantan, P.~Shyam, G.~Sastry, A.~Askell, S.~Agarwal, A.~Herbert{-}Voss, G.~Krueger, T.~Henighan, R.~Child, A.~Ramesh, D.~M. Ziegler, J.~Wu, C.~Winter, C.~Hesse, M.~Chen, E.~Sigler, M.~Litwin, S.~Gray, B.~Chess, J.~Clark, C.~Berner, S.~McCandlish, A.~Radford, I.~Sutskever, and D.~Amodei.
\newblock Language models are few-shot learners.
\newblock \emph{CoRR}, abs/2005.14165, 2020.
\newblock URL \url{https://arxiv.org/abs/2005.14165}.

\bibitem[Gao et~al.(2023)Gao, Xiong, Gao, Jia, Pan, Bi, Dai, Sun, and Wang]{rag_survey}
Y.~Gao, Y.~Xiong, X.~Gao, K.~Jia, J.~Pan, Y.~Bi, Y.~Dai, J.~Sun, and H.~Wang.
\newblock Retrieval-augmented generation for large language models: A survey.
\newblock \emph{arXiv preprint arXiv:2312.10997}, 2023.

\bibitem[Gu et~al.(2023)Gu, Dong, Wei, and Huang]{PICL}
Y.~Gu, L.~Dong, F.~Wei, and M.~Huang.
\newblock Pre-training to learn in context.
\newblock \emph{arXiv preprint arXiv:2305.09137}, 2023.

\bibitem[Sridhar et~al.(2025)Sridhar, Dutta, Jayaraman, and Lee]{kaustubh_regent_iclr}
K.~Sridhar, S.~Dutta, D.~Jayaraman, and I.~Lee.
\newblock {REGENT}: A retrieval-augmented generalist agent that can act in-context in new environments.
\newblock In \emph{The Thirteenth International Conference on Learning Representations}, 2025.
\newblock URL \url{https://openreview.net/forum?id=NxyfSW6mLK}.

\bibitem[Goyal et~al.(2023)Goyal, Kumar, Garg, Kolter, and Raghunathan]{finetune_like_you_pretrain}
S.~Goyal, A.~Kumar, S.~Garg, Z.~Kolter, and A.~Raghunathan.
\newblock Finetune like you pretrain: Improved finetuning of zero-shot vision models.
\newblock In \emph{Proceedings of the IEEE/CVF Conference on Computer Vision and Pattern Recognition}, pages 19338--19347, 2023.

\bibitem[Reed et~al.(2022)Reed, Zolna, Parisotto, Colmenarejo, Novikov, Barth-Maron, Gimenez, Sulsky, Kay, Springenberg, et~al.]{gato}
S.~Reed, K.~Zolna, E.~Parisotto, S.~G. Colmenarejo, A.~Novikov, G.~Barth-Maron, M.~Gimenez, Y.~Sulsky, J.~Kay, J.~T. Springenberg, et~al.
\newblock A generalist agent.
\newblock \emph{arXiv preprint arXiv:2205.06175}, 2022.

\bibitem[Gallou{\'e}dec et~al.(2024)Gallou{\'e}dec, Beeching, Romac, and Dellandr{\'e}a]{JAT}
Q.~Gallou{\'e}dec, E.~Beeching, C.~Romac, and E.~Dellandr{\'e}a.
\newblock Jack of all trades, master of some, a multi-purpose transformer agent.
\newblock \emph{arXiv preprint arXiv:2402.09844}, 2024.

\bibitem[Raparthy et~al.(2023)Raparthy, Hambro, Kirk, Henaff, and Raileanu]{mtt}
S.~C. Raparthy, E.~Hambro, R.~Kirk, M.~Henaff, and R.~Raileanu.
\newblock Generalization to new sequential decision making tasks with in-context learning.
\newblock \emph{arXiv preprint arXiv:2312.03801}, 2023.

\bibitem[Bruce et~al.(2024)Bruce, Dennis, Edwards, Parker-Holder, Shi, Hughes, Lai, Mavalankar, Steigerwald, Apps, et~al.]{genie}
J.~Bruce, M.~D. Dennis, A.~Edwards, J.~Parker-Holder, Y.~Shi, E.~Hughes, M.~Lai, A.~Mavalankar, R.~Steigerwald, C.~Apps, et~al.
\newblock Genie: Generative interactive environments.
\newblock In \emph{Forty-first International Conference on Machine Learning}, 2024.

\bibitem[Fu et~al.(2024)Fu, Huang, Datta, Chen, Panitch, Liu, Li, and Goldberg]{icrt}
L.~Fu, H.~Huang, G.~Datta, L.~Y. Chen, W.~C.-H. Panitch, F.~Liu, H.~Li, and K.~Goldberg.
\newblock In-context imitation learning via next-token prediction.
\newblock \emph{arXiv preprint arXiv:2408.15980}, 2024.

\bibitem[Di~Palo and Johns(2024)]{kat}
N.~Di~Palo and E.~Johns.
\newblock Keypoint action tokens enable in-context imitation learning in robotics.
\newblock \emph{arXiv preprint arXiv:2403.19578}, 2024.

\bibitem[Papagiannis et~al.(2024)Papagiannis, Di~Palo, Vitiello, and Johns]{r+x}
G.~Papagiannis, N.~Di~Palo, P.~Vitiello, and E.~Johns.
\newblock R+ x: Retrieval and execution from everyday human videos.
\newblock \emph{arXiv preprint arXiv:2407.12957}, 2024.

\bibitem[Yin et~al.(2024)Yin, Wang, Sharma, Niu, Darrell, and Herzig]{roboprompt}
Y.~Yin, Z.~Wang, Y.~Sharma, D.~Niu, T.~Darrell, and R.~Herzig.
\newblock In-context learning enables robot action prediction in llms.
\newblock \emph{arXiv preprint arXiv:2410.12782}, 2024.

\bibitem[Mirchandani et~al.(2023)Mirchandani, Xia, Florence, Ichter, Driess, Arenas, Rao, Sadigh, and Zeng]{llm_pattern_machines}
S.~Mirchandani, F.~Xia, P.~Florence, B.~Ichter, D.~Driess, M.~G. Arenas, K.~Rao, D.~Sadigh, and A.~Zeng.
\newblock Large language models as general pattern machines.
\newblock \emph{arXiv preprint arXiv:2307.04721}, 2023.

\bibitem[Ma et~al.(2023)Ma, Liang, Wang, Huang, Bastani, Jayaraman, Zhu, Fan, and Anandkumar]{Eureka}
Y.~J. Ma, W.~Liang, G.~Wang, D.-A. Huang, O.~Bastani, D.~Jayaraman, Y.~Zhu, L.~Fan, and A.~Anandkumar.
\newblock Eureka: Human-level reward design via coding large language models.
\newblock \emph{arXiv preprint arXiv:2310.12931}, 2023.

\bibitem[Ma et~al.(2024)Ma, Hejna, Fu, Shah, Liang, Xu, Kirmani, Xu, Driess, Xiao, et~al.]{GVL}
Y.~J. Ma, J.~Hejna, C.~Fu, D.~Shah, J.~Liang, Z.~Xu, S.~Kirmani, P.~Xu, D.~Driess, T.~Xiao, et~al.
\newblock Vision language models are in-context value learners.
\newblock In \emph{The Thirteenth International Conference on Learning Representations}, 2024.

\bibitem[Khazatsky et~al.(2024)Khazatsky, Pertsch, Nair, Balakrishna, Dasari, Karamcheti, Nasiriany, Srirama, Chen, Ellis, et~al.]{droid}
A.~Khazatsky, K.~Pertsch, S.~Nair, A.~Balakrishna, S.~Dasari, S.~Karamcheti, S.~Nasiriany, M.~K. Srirama, L.~Y. Chen, K.~Ellis, et~al.
\newblock Droid: A large-scale in-the-wild robot manipulation dataset.
\newblock \emph{arXiv preprint arXiv:2403.12945}, 2024.

\bibitem[Oquab et~al.(2023)Oquab, Darcet, Moutakanni, Vo, Szafraniec, Khalidov, Fernandez, Haziza, Massa, El-Nouby, et~al.]{dino-v2}
M.~Oquab, T.~Darcet, T.~Moutakanni, H.~Vo, M.~Szafraniec, V.~Khalidov, P.~Fernandez, D.~Haziza, F.~Massa, A.~El-Nouby, et~al.
\newblock Dinov2: Learning robust visual features without supervision.
\newblock \emph{arXiv preprint arXiv:2304.07193}, 2023.

\bibitem[Borgeaud et~al.(2022)Borgeaud, Mensch, Hoffmann, Cai, Rutherford, Millican, Van Den~Driessche, Lespiau, Damoc, Clark, et~al.]{borgeaud2022RETRO}
S.~Borgeaud, A.~Mensch, J.~Hoffmann, T.~Cai, E.~Rutherford, K.~Millican, G.~B. Van Den~Driessche, J.-B. Lespiau, B.~Damoc, A.~Clark, et~al.
\newblock Improving language models by retrieving from trillions of tokens.
\newblock In \emph{International conference on machine learning}, pages 2206--2240. PMLR, 2022.

\bibitem[Sridhar et~al.(2024)Sridhar, Dutta, Jayaraman, Weimer, and Lee]{kaustubh_mcnn_iclr}
K.~Sridhar, S.~Dutta, D.~Jayaraman, J.~Weimer, and I.~Lee.
\newblock Memory-consistent neural networks for imitation learning.
\newblock In \emph{The Twelfth International Conference on Learning Representations}, 2024.
\newblock URL \url{https://openreview.net/forum?id=R3Tf7LDdX4}.

\bibitem[Pari et~al.(2021)Pari, Shafiullah, Arunachalam, and Pinto]{pari2021VINN}
J.~Pari, N.~M. Shafiullah, S.~P. Arunachalam, and L.~Pinto.
\newblock The surprising effectiveness of representation learning for visual imitation.
\newblock \emph{arXiv preprint arXiv:2112.01511}, 2021.

\bibitem[Lin et~al.(2024)Lin, Cui, Xie, Hua, and Sadigh]{dorsa_flow_retrieval}
L.-H. Lin, Y.~Cui, A.~Xie, T.~Hua, and D.~Sadigh.
\newblock Flowretrieval: Flow-guided data retrieval for few-shot imitation learning.
\newblock \emph{arXiv preprint arXiv:2408.16944}, 2024.

\bibitem[Du et~al.(2023)Du, Nair, Sadigh, and Finn]{dorsa_behavior_retrieval}
M.~Du, S.~Nair, D.~Sadigh, and C.~Finn.
\newblock Behavior retrieval: Few-shot imitation learning by querying unlabeled datasets.
\newblock \emph{arXiv preprint arXiv:2304.08742}, 2023.

\bibitem[Beyer et~al.(2024)Beyer, Steiner, Pinto, Kolesnikov, Wang, Salz, Neumann, Alabdulmohsin, Tschannen, Bugliarello, et~al.]{paligemma}
L.~Beyer, A.~Steiner, A.~S. Pinto, A.~Kolesnikov, X.~Wang, D.~Salz, M.~Neumann, I.~Alabdulmohsin, M.~Tschannen, E.~Bugliarello, et~al.
\newblock Paligemma: A versatile 3b vlm for transfer.
\newblock \emph{arXiv preprint arXiv:2407.07726}, 2024.

\bibitem[Zhai et~al.(2023)Zhai, Mustafa, Kolesnikov, and Beyer]{siglip}
X.~Zhai, B.~Mustafa, A.~Kolesnikov, and L.~Beyer.
\newblock Sigmoid loss for language image pre-training.
\newblock In \emph{Proceedings of the IEEE/CVF international conference on computer vision}, pages 11975--11986, 2023.

\bibitem[Team et~al.(2024)Team, Mesnard, Hardin, Dadashi, Bhupatiraju, Pathak, Sifre, Rivi{\`e}re, Kale, Love, et~al.]{gemma}
G.~Team, T.~Mesnard, C.~Hardin, R.~Dadashi, S.~Bhupatiraju, S.~Pathak, L.~Sifre, M.~Rivi{\`e}re, M.~S. Kale, J.~Love, et~al.
\newblock Gemma: Open models based on gemini research and technology.
\newblock \emph{arXiv preprint arXiv:2403.08295}, 2024.

\end{thebibliography}

%%%%%%%%%%%%%%%%%%%%%%%%%%%
\clearpage
\appendix
\section*{Appendix}
% \section{Supplementary Video (strict 100 MB limit, suggested 3 mins) with overview of work}

\section{Additional Related Work}
We discuss some other relevant work below.

\textbf{Retrieval in robotics: }Recent work on retrieval in robot learning has created surprisingly adaptive and reliable policies \citep{kaustubh_mcnn_iclr, pari2021VINN, dorsa_flow_retrieval, dorsa_behavior_retrieval}. Early work in VINN \citep{pari2021VINN} and recent work in MCNN \citep{kaustubh_mcnn_iclr} demonstrates the capabilities of retrieval based on image embeddings to improve behavior cloning policies. Recent work in behavior \citep{dorsa_behavior_retrieval} and flow retrieval \citep{dorsa_flow_retrieval} demonstrate the ability of retrieval to augment task-specific data with a subset of offline data to further improve behavior cloning policies. While \RICL{} builds on the former use of retrieval, we believe future work that includes the latter use of retrieval could further improve \RICL{}-VLAs that are finetuned on task-specific demonstrations.

\section{Additional Background Information}
\label{app:more_background_regentic_tuning}
We provide additional background information below.

\textbf{$\pi_0$-FAST: }It was created by finetuning the PaliGemma Vision-Language model (VLM) \citep{paligemma} on a (unknown but) large number of robot trajectories from different embodiments. The PaliGemma VLM is a combination of a SigLIP image encoder \citep{siglip} and the 3B parameters Gemma large language model (LLM) \citep{gemma}. In particular, it differentiates itself from the similar $\pi_0$ diffusion-based VLA \citep{pi0_diffusion} with the use of the FAST tokenizer \citep{pi0_fast} to bring action tokens into the text token space for simple autoregressive prediction.

\textbf{$\pi_0$-FAST-DROID: }$\pi_0$-FAST-DROID was created by finetuning $\pi_0-FAST$ on the DROID dataset \citep{droid}. The DROID dataset was collected with the Franka DROID platform--which consists of a Franka arm on a movable platform with cameras on the left, right, and wrist--in-the-wild, across universities and scenes and with a large variety of objects. Each episode in the DROID dataset also has proprioceptive states, actions, and language annotations. In this work, we use the DROID platform (see Figures \ref{fig:franka_droid_setup} and \ref{fig:regentic-pi0-arch}) for all results and hence, we will refer to this model frequently in the rest of the paper.

\textbf{REGENT }\citep{kaustubh_regent_iclr}, as discussed above, trains a transformer based multi-task policy on sequences of query and retrieved state-action pairs, \textit{from scratch}, on 145 simulated robotics tasks and games. Given a few demonstrations in a held out simulated environment or game, \Regent{} retrieves state-action pairs from these demonstrations, throws it into its context, and plays the held out game. %\jd{feels repetitive, just merge this with the discussion of RAG + ICL? Or should it instead be in method to avoid the method sounding like it ``just'' combines A and B?} \ks{this is not even needed because as DInesh said it is repetitive}

\section{Additional Info on Franka DROID Setup, \RICL{}, and \RICL{}-\texorpdfstring{$\pi_0$}{Pi0}-FAST-DROID}
\label{app:more_details_regentic_tuning}

\begin{figure}[t]
    \centering
    \includegraphics[width=0.49\linewidth]{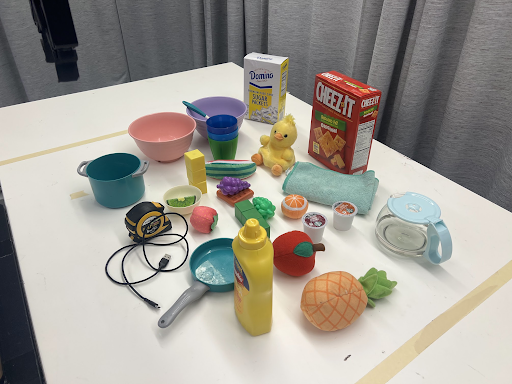} \hspace{0.5cm}
    \includegraphics[width=0.278\linewidth]{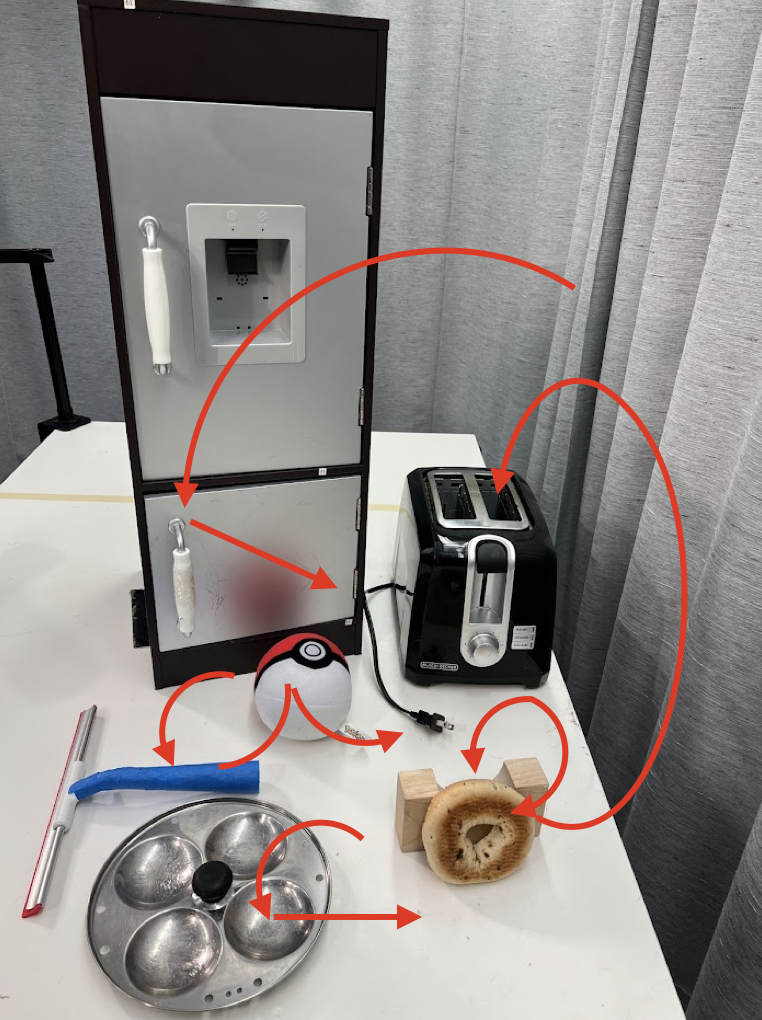}
    \caption{\small \textbf{[LEFT]} Objects used during \RICL{}, either as the primary object in the task or a distractor, to create \RICL{}-$\pi_0$-FAST-DROID. \textbf{[RIGHT]} Unseen objects and depictions of novel motions in red arrows. Unseen tasks also include distractors from the objects used during training.}
    \label{fig:training_and_unseen_objects}
\end{figure}

\begin{figure}[t]
    \centering
    \includegraphics[width=\linewidth]{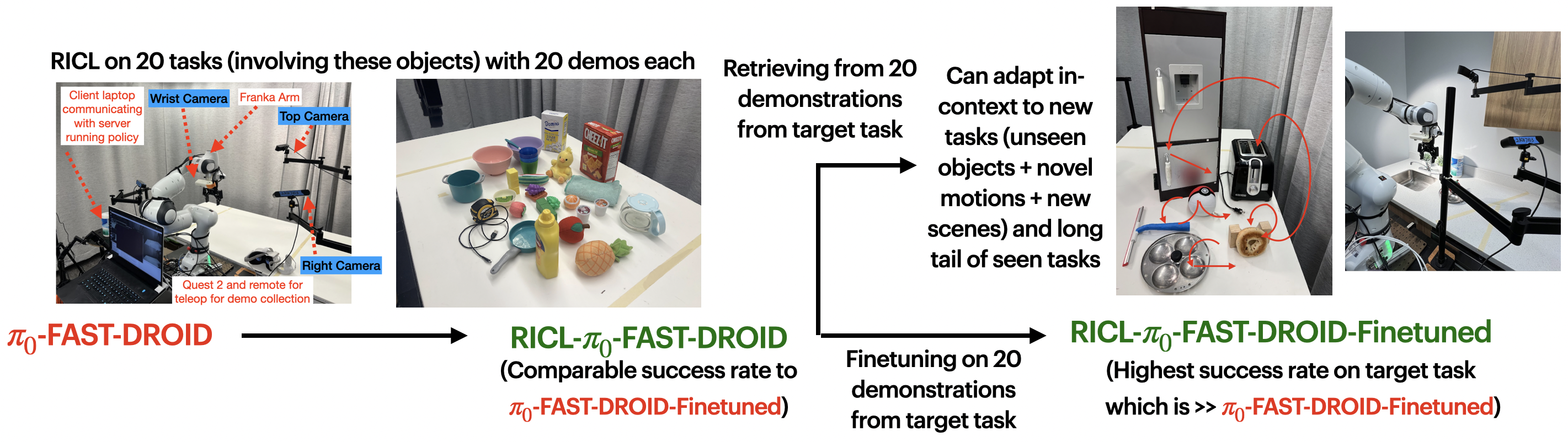}
    \caption{\small Overview of \RICL{}.}
    \label{fig:overview}
\end{figure}

\textbf{Exact list of tasks for \RICL{}: } We collect 20 demonstrations in each of the following 20 tasks (for a total of about 400 demonstrations):
\begin{itemize}
    \item 5x of "move <object> to the right" where object in [apple, orange, strawberry, the coffee pod, the cup],
    \item 5x of “move <object> to the left” where object in [box, cup, duck, pan, pot], 
    \item 6x of “pick up <object> and put it in the bowl” where object in [the block, the cloth, the grapes on the waffle, the orange, the pineapple, the watermelon slice], 
    \item 1x of “pour the contents of the mug into the bowl”
    \item 1x of “pick up the spoon in the bowl and put it in the other bowl”
    \item 1x of “put the cable in the bowl”
    \item 1x of “move the watermelon slice from one bowl to the other bowl”
\end{itemize} 
the pictures of these objects and distractors used can be seen in Figure \ref{fig:training_and_unseen_objects}. We also depict the unseen objects and novel motions in the same Figure. We note that we collect data at an action frequency of 15 Hz. 

\textbf{Observation and action space: }The proprioceptive observation consists of 7 joint angle and gripper position. The action chunks are of shape (15, 8) where 15 corresponds to the prediction horizon (which we empirically found to have better performance than 10). Each action has 7 joint velocities and 1 gripper position. All images are resized with padding to 224$\times$224.

% write hypers in appendix (regent, regent-fine, pi0-finet,dp)

\textbf{More details and key hyperparameters for \RICL{} on priming demonstrations to create \RICL{}-$\pi_0$-FAST-DROID: }We use four retrieved groups in our context and set $\lambda=10$. 
For all other hyperparameters we build on top of the openpi repository\footnote{\url{https://github.com/Physical-Intelligence/openpi}} and make the following key changes: using action chunks of 15 steps, training for three epochs, with a CosineDecaySchedule with 300 warmup steps, 2.5e-5 peak learning rate, 3000 decay steps, and 2.5e-6 decay learning rate. We also use a batch size of 16 and two A100 GPUs. Our detailed codebase can be found in our website.
% During training, we perform the following data augmentations, following the openpi repository, on the images in the training sequences: randomcrop to 95\% of 224 in both width and height, resizing back to 224$\times$224, rotating by a random angle in $(-5, 5)$ degrees, with a random color jitter (with only the color jitter for the wrist image).

\textbf{Key hyperparameters for further finetuning \RICL{}-$\pi_0$-FAST-DROID on each task's 20 demonstrations: }We use a recipe similar to \RICL{} but only finetune for a 1000 steps, and with the same learning rate scheduler but only for 1000 decay steps with 50 warmup steps.

\textbf{Key hyperparameters for further $\pi_0$ finetuning on each task's 20 demonstrations: }We use the same number of train setps, decay steps and warmup steps in the optimizer as above but otherwise use the recommended pi0 fionetuning recipe in the openpi repository.

\textbf{Key hyperparameters at deployment: }For \RICL{}-$\pi_0$-FAST-DROID, we execute 3 of the predicted 15 actions in the action chunk until a gripping action is predicted in the last action of an action chunk, at which point we switch to using 8 of the predicted 15 actions in an action chunk. Similarly, for \RICL{}-$\pi_0$-FAST-DROID-finetuned, we execute 3 of the predicted 15 until a gripping action is predicted, at which point we switch to using 5 of the predicted 15.

\textbf{Discussion on runtime of \RICL{}: }The retrieval step occurs in less than a millisecond with the help of an existing search index library (FAISS). The 4x increase in number of tokens only results in RICL-pi0-FAST taking about 1.33x the time of pi0-FAST in a rollout. This can be observed in the videos on our website where pi0 is at 6x real time and RICL-pi0 at about 8x real time.

\textbf{Discussion on \RICL{}'s performance when retrieval demonstration data is different from test conditions: }During our experiments, minor changes did occur in the camera viewpoint and the condition of the scene (positions of distractors, tape on the tables, etc), and RICL did handle these minor changes. However, RICL cannot currently handle more significant changes to the camera viewpoint or scene. We believe that if RICL is trained in the priming phase with such changes then it would be more robust to such changes.

\section{Additional Results}
\label{app:additional_results_regentic_tuning}
% \newpage 
% \input{regentic_tuning/tables/results_table}
% \clearpage  
% \input{regentic_tuning/tables/ablations_table}

\begin{figure}[t]
    \centering
    \begin{subfigure}{\textwidth}
        \centering
        \fcolorbox{red}{white}{\includegraphics[width=0.35\linewidth]{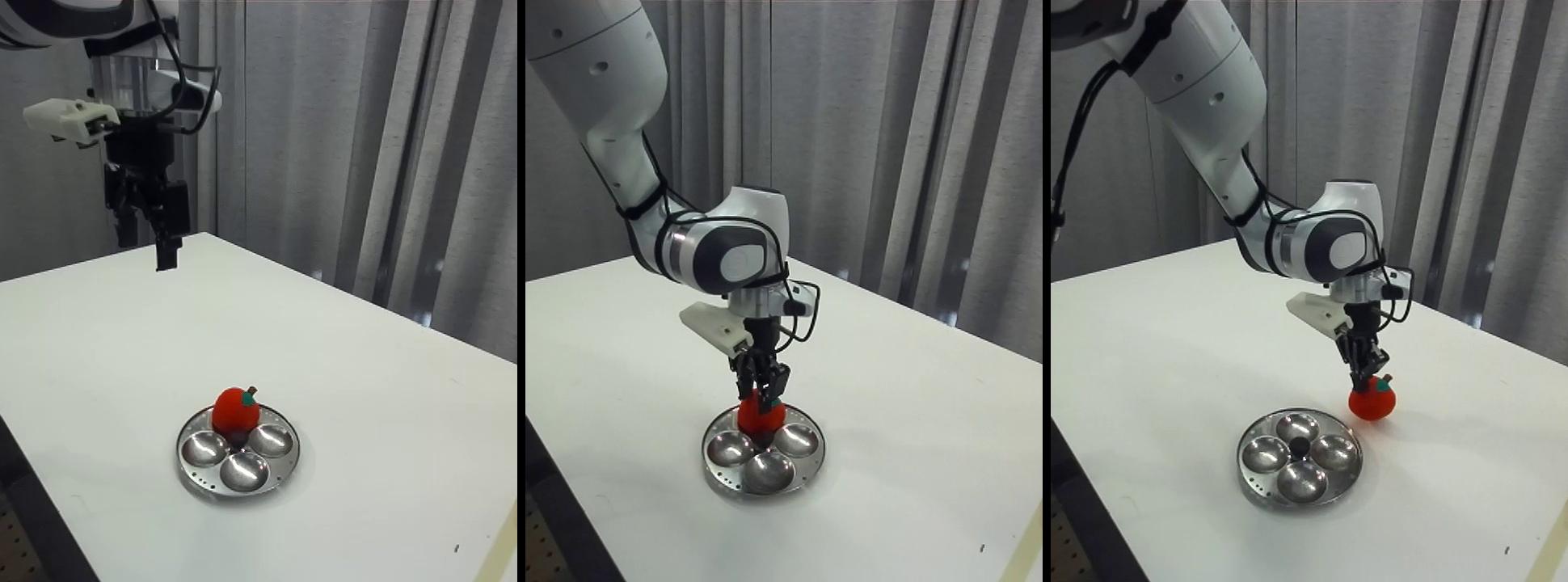}} \hspace{1em}
        \fcolorbox{green}{white}{\includegraphics[width=0.47\linewidth]{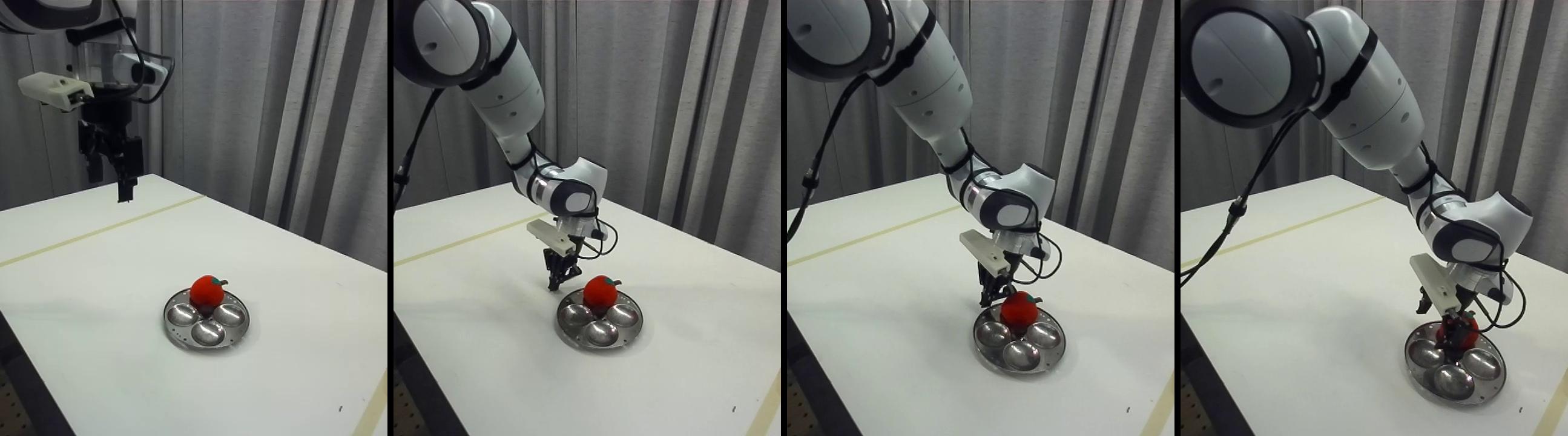}}
        \vspace{-0.5em}
        \caption{\scriptsize Task: "move the idli plate to the right". $\pi_0$-FAST-DROID \textbf{[L]} moves the apple sitting on the plate (language grounding issue). \RICL{}-$\pi_0$-FAST-DROID \textbf{[R]} moves the unseen object (idli plate) with depressions that require an unfamiliar grasp, only with RAG and ICL. Further, it does so with elicited latent actions not in the retrieval data (see Section \ref{sec:experiments_regentic_tuning}).}
    \end{subfigure}
    \begin{subfigure}{\textwidth}
        \centering
        \fcolorbox{red}{white}{\includegraphics[width=0.35\linewidth]{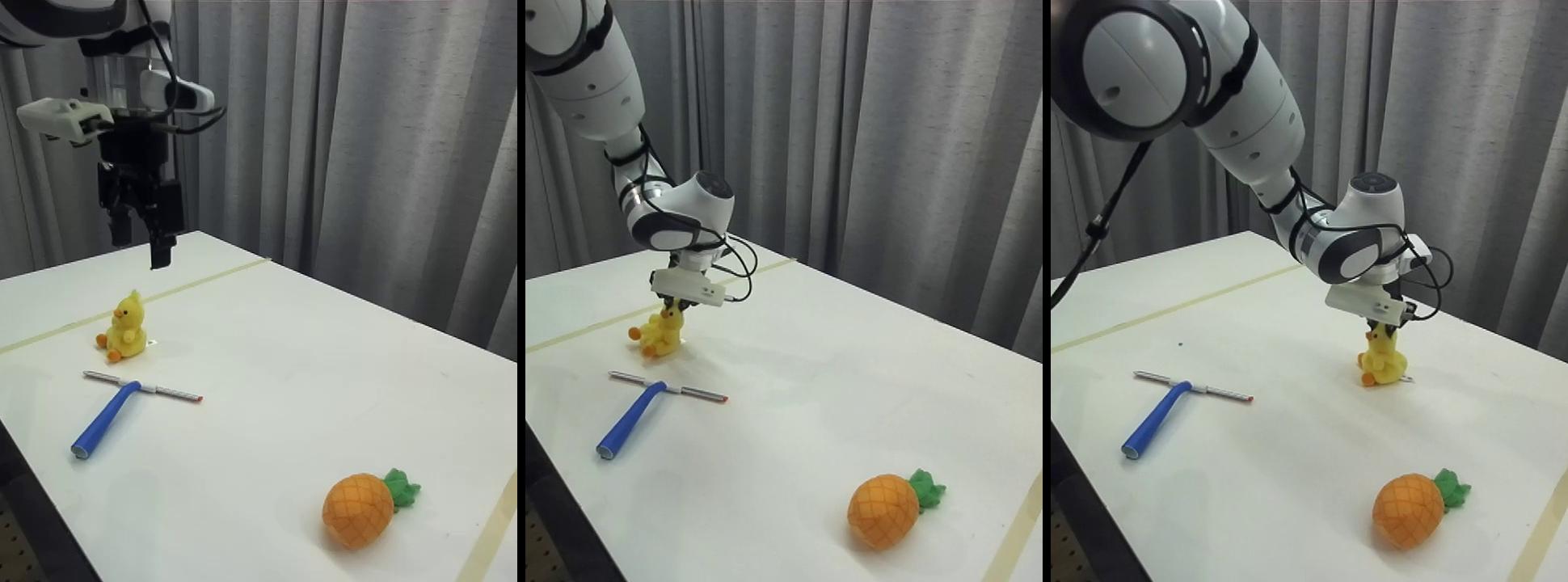}} \hspace{1em}
        \fcolorbox{green}{white}{\includegraphics[width=0.47\linewidth]{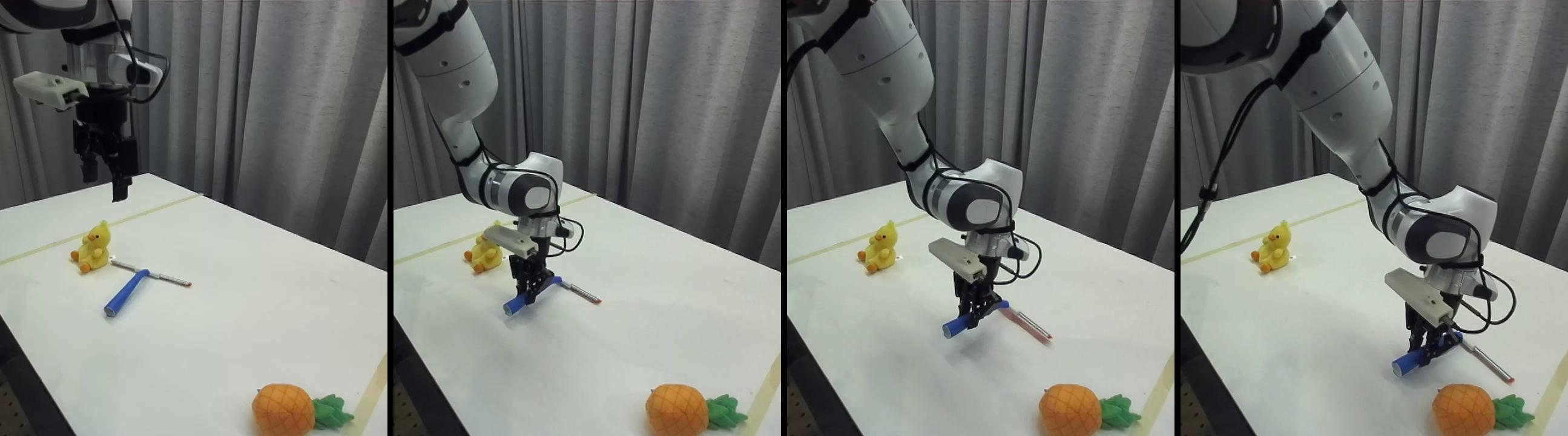}}
        \vspace{-0.5em}
        \caption{\scriptsize Task: "move the squeegee to the right and try to drag it". $\pi_0$-FAST-DROID \textbf{[L]} moves the distractor (duck) instead (language grounding issue). In other rollouts, it cannot figure out the grasp or the motion (adaptation issue). \RICL{}-$\pi_0$-FAST-DROID \textbf{[R]} moves the unseen object (squeegee) in a novel motion of partly lifting and partly dragging, only with RAG and ICL}
    \end{subfigure}
    \caption{\small Additional qualitative comparisons between $\pi_0$-FAST-DROID \textbf{[L]} and \RICL{}-$\pi_0$-FAST-DROID \textbf{[R]}, with 20 task specific demonstrations for RAG and ICL, on new tasks, including novel objects, motions, and scenes.}
    \label{fig:qualitative_results_appendix}
\end{figure}

\begin{figure}[t]
    \centering
    \begin{subfigure}{\textwidth}
        \centering
        \includegraphics[width=0.75\linewidth]{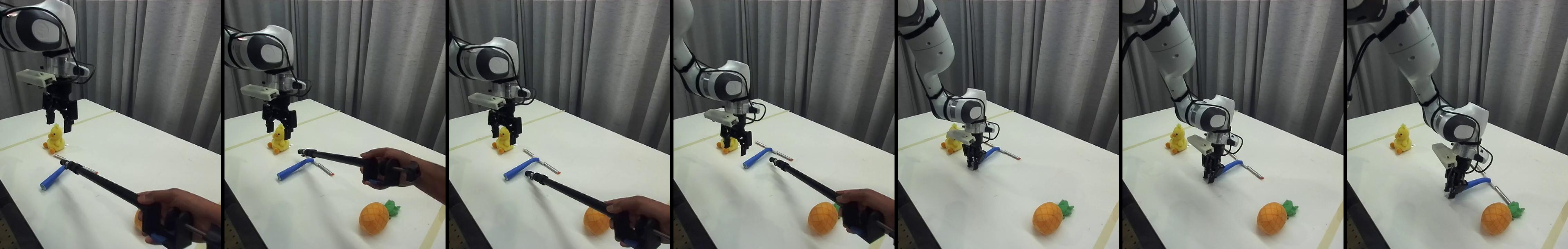}
        \vspace{-0.5em}
        \caption{\scriptsize Task: "move the squeegee to the right and try to drag it"}
    \end{subfigure}
    \begin{subfigure}{\textwidth}
        \centering
        \includegraphics[width=0.75\linewidth]{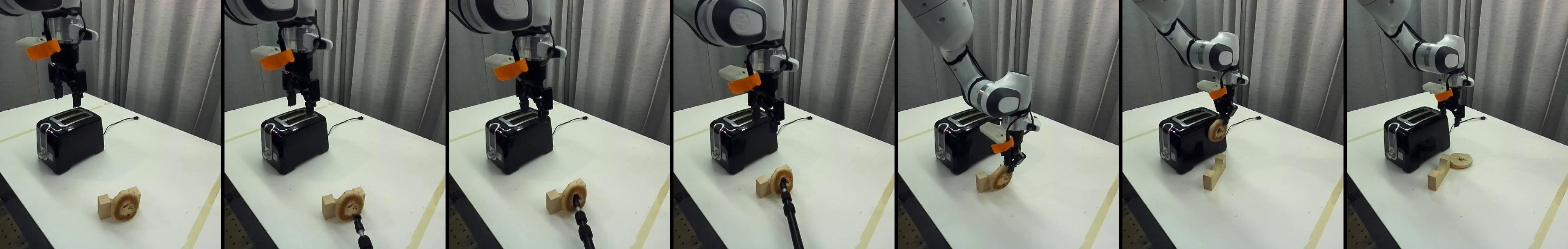}
        \vspace{-0.5em}
        \caption{\scriptsize Task: "pick up the bagel and put it in the toaster"}
    \end{subfigure}
    \begin{subfigure}{\textwidth}
        \centering
        \includegraphics[width=0.75\linewidth]{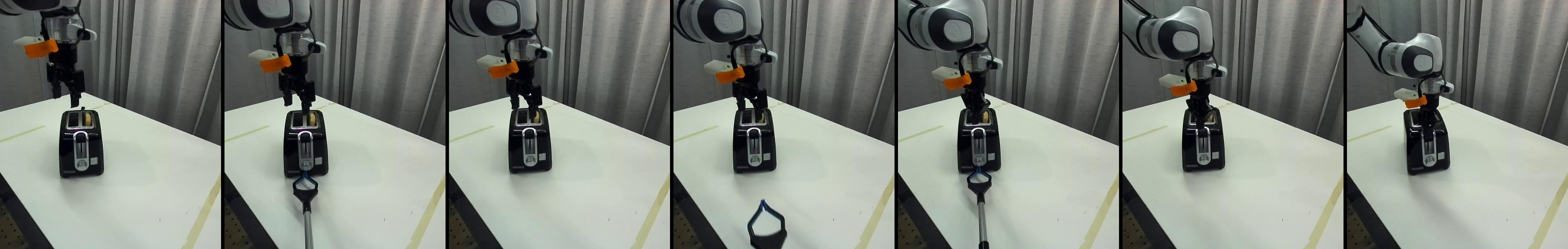}
        \vspace{-0.5em}
        \caption{\scriptsize Task: "push the lever on the toaster"}
    \end{subfigure}
    \caption{\small Additional qualitative visualizations of the reactivity and robustness of \RICL{}-$\pi_0$-FAST-DROID-finetuned on 20 task-specific demonstrations in a dynamic test rollout. In the above, a human randomly perturbs and displaces the primary object during the test rollout.}
    \label{fig:qualitative_reliability_results_appendix}
\end{figure}

We provide additional qualitative comparisons between $\pi_0$-FAST-DROID and \RICL{}-$\pi_0$-FAST-DROID in Figure \ref{fig:qualitative_results_appendix}. We also provide additional qualitative visualizations of the reactivity and robustness of \RICL{}-$\pi_0$-FAST-DROID-Finetuned in Figure \ref{fig:qualitative_reliability_results_appendix}.

\begin{figure}[t]
    \centering
    \begin{subfigure}{\textwidth}
        \centering
        \includegraphics[width=\linewidth]{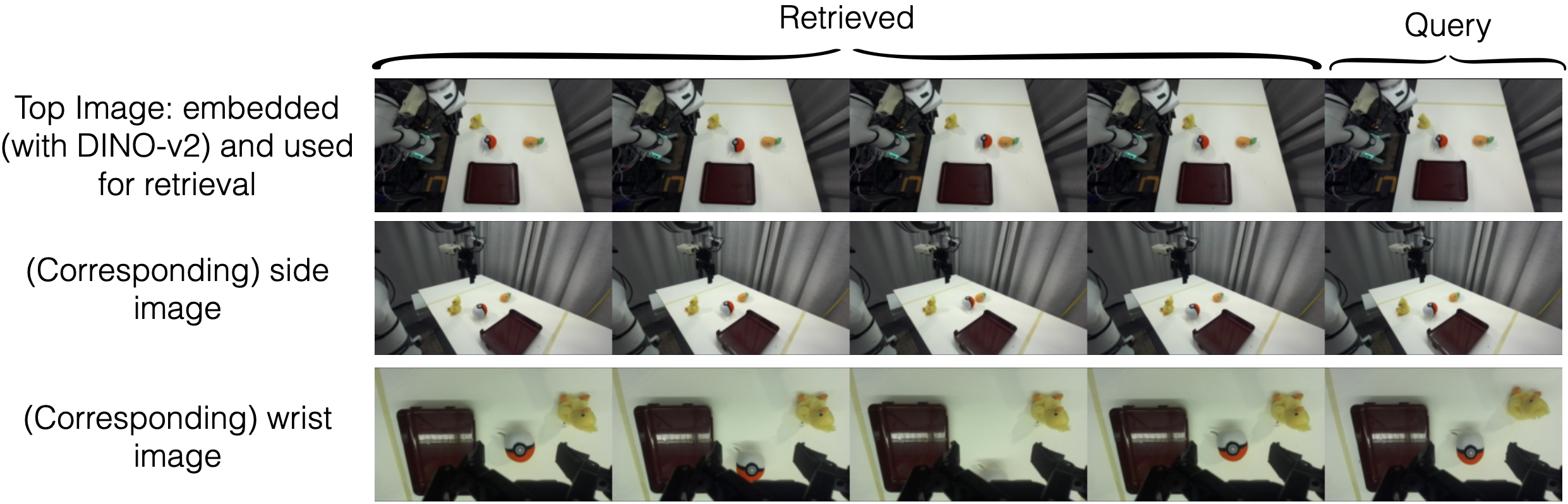}
        \caption{First state of the test rollout}
    \end{subfigure}
    \begin{subfigure}{\textwidth}
        \centering
        \includegraphics[width=\linewidth]{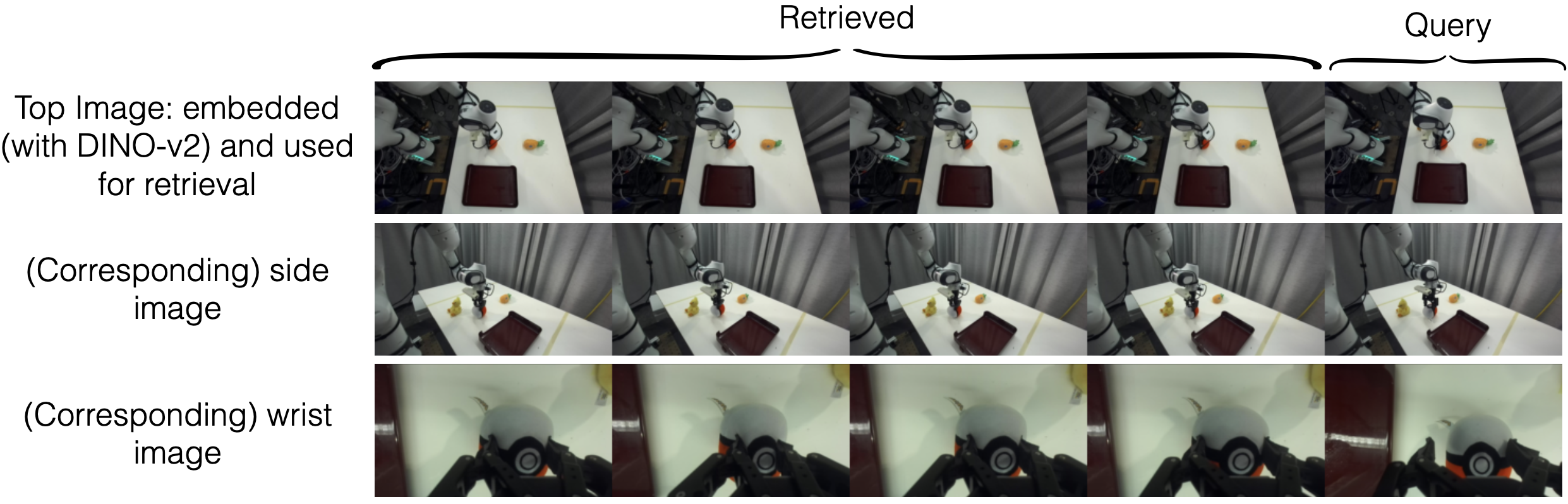}
        \caption{An intermediate state of the test rollout when the Franka arm starts to grasp the pokeball}
    \end{subfigure}
    \begin{subfigure}{\textwidth}
        \centering
        \includegraphics[width=\linewidth]{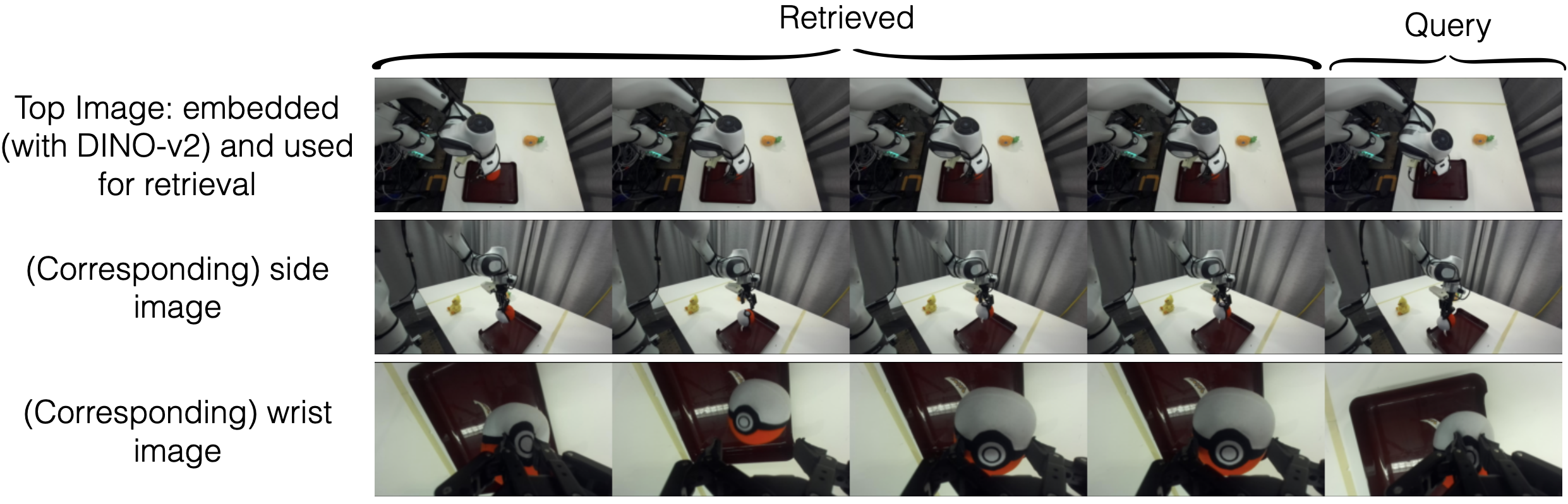}
        \caption{Last state of the test rollout at which an action chunk was predicted}
    \end{subfigure}
    \caption{\small Visualization of Query and Retrieved images at three states during a \RICL{}-$\pi_0$-FAST-DROID test rollout on the \texttt{pokeball} task.}
    \label{fig:retrieved_query_viz_pokeball}
\end{figure}

\textbf{Understanding the quality of retrieval (and hence the quality of the context): }To understand the quality of retrieval, we depict retrieved top images and their corresponding side and wrist images for key states in Figure \ref{fig:retrieved_query_viz_pokeball} 
% \jd{this is a lot of space, for not much takeaway. Can you point to something about the images that is particularly compelling / surprising? Otherwise prime candidate for moving to appendix.} 
for the poke ball task. We find that our simple retrieval mechanism can find similar states with close initial positions and 
% we believe \jd{kind of a weird word here, is this not testable?} 
\RICL{}-$\pi_0$-FAST-DROID interpolates amongst these similar states and actions to predict a suitable action.

\end{document}